\documentclass[10pt,twocolumn,letterpaper]{article}

\usepackage{cvpr}              
\definecolor{cvprblue}{rgb}{0.21,0.49,0.74}
\usepackage[pagebackref,breaklinks,colorlinks,allcolors=cvprblue]{hyperref}
\usepackage{colortbl}
\usepackage{xcolor}
\usepackage{booktabs}
\usepackage{multirow}
\usepackage{boldline} 
\usepackage{needspace}
\definecolor{skyblue1}{RGB}{115, 174, 218} 
\definecolor{skyblue2}{RGB}{210, 231, 247} 

\def\confName{CVPR}
\def\confYear{2027}

\title{RQUL-UIE: Revitalizing Quality-Unstable Labels for Underwater Image Enhancement via Quality-Aware Supervision Allocation}

\author{
Haochen Hu$^{1,2}$ \thanks{Part of this work was conducted during internship at Seahi Inc.} \quad
Yanrui Bin$^{1}$ \quad
Qihang Peng$^{1}$ \quad
Chih-yung Wen$^{1}$ \quad
Bing Wang$^{1}$ \thanks{Corresponding author. Email: bingwang@connect.polyu.hk.} \\
$^{1}$The Hong Kong Polytechnic University \qquad $^{2}$Seahi Inc.\\
}

\begin{document}
\maketitle
\begin{abstract}
Learning-based underwater image enhancement (UIE) often relies on paired datasets whose reference images vary substantially in quality. Treating these references as equally reliable targets limits enhancement performance, while filtering pairs with modest corrections reduces the available supervision. We propose RQUL-UIE, a framework that uses estimated input-to-reference correction to determine where each reference supervises a shared recurrent enhancer. Specifically, level-wise latent space enhancement (LLSE) assigns smaller estimated corrections to shallower supervision stages and larger corrections to deeper stages, allowing all pairs to contribute without additional references or quality annotations. To compensate for detail loss caused by the variational Autoencoder (VAE)'s lossy compression, texture refinement via frequency modulation (TRFDM) draws on cues retained in the raw image. Evaluated against 12 baselines, the proposal achieves the best overall performance across four metrics on the three evaluation collections. Qualitative comparisons further show improved color balance and clearer structures in degraded underwater scenes. Project link\url{https://github.com/Haru2022/RQUL-UIE}.
\end{abstract}
    
\section{Introduction}
\label{sec:intro}

\begin{figure}[htbp]
    \centering
    \includegraphics[width=1\linewidth]{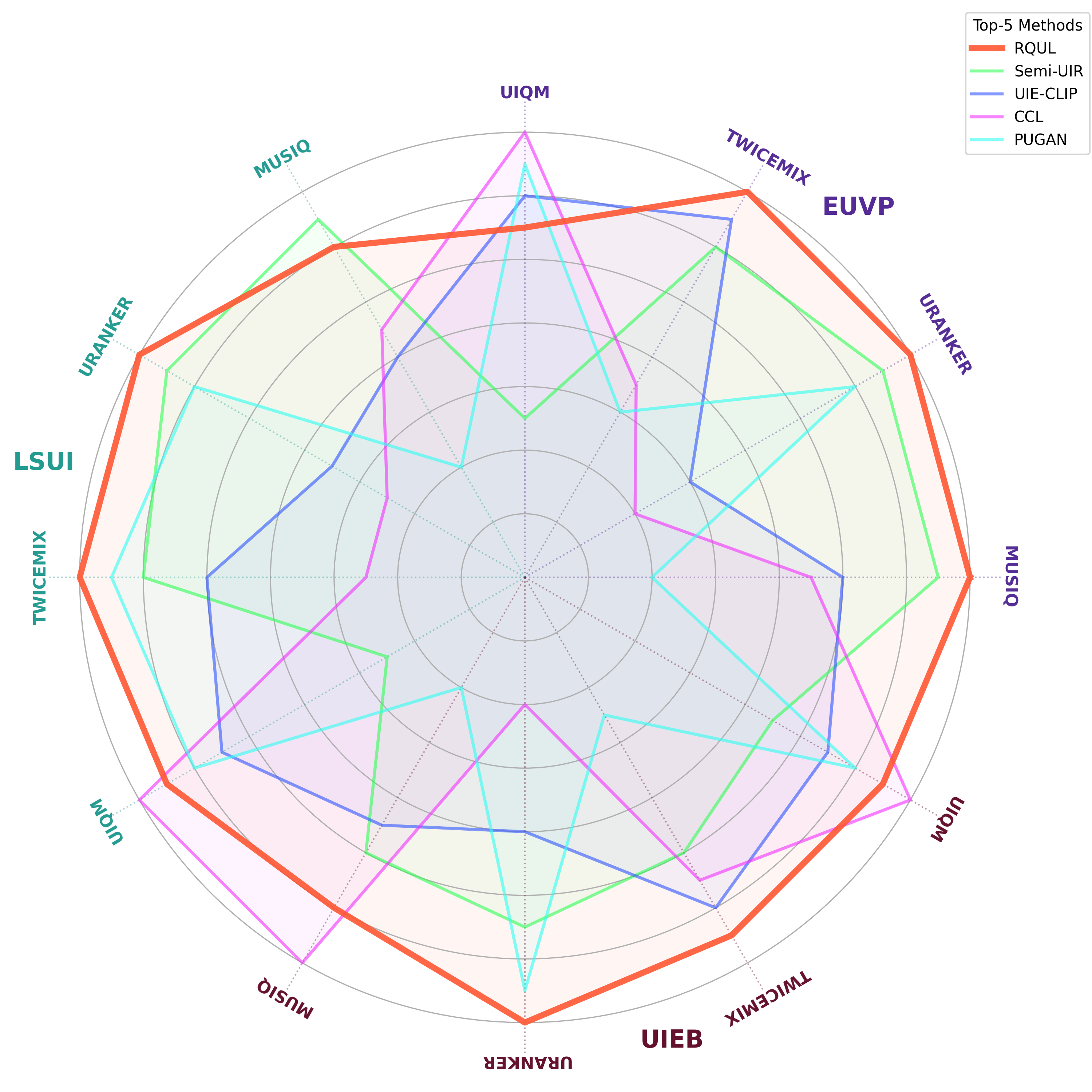}
    \caption{Comparison of the top-five methods on LSUI \cite{peng2023ushape}, EUVP \cite{islam2020euvp}, and UIEB \cite{li2019uieb}, evaluated using MUSIQ \cite{ke2021musiq}, URANKER \cite{guo2023uranker}, TWICEMIX \cite{fu2022twicemix}, and UIQM \cite{panetta2015uiqm}. RQUL-UIE achieves the best overall average performance. UIEB is the training source collection for RQUL-UIE.}
    \label{fig:radar_overview}
\end{figure}

Underwater imaging supports autonomous underwater vehicles, infrastructure inspection, and marine research. However, absorption and scattering effects introduced by the water medium cause color casts and contrast reduction \cite{gonzalez2023survey}, which heavily impair visual information and downstream vision-based perception systems. To address these degradations, underwater image enhancement (UIE) has attracted increasing attention in recent years.

\begin{figure*}[htbp]
    \centering
    \includegraphics[width=\linewidth]{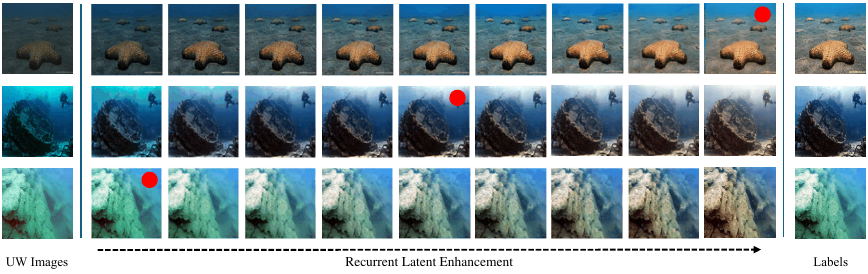}
    \caption{Supervision depth assignment by estimated correction extent. Within the proposed LLSE, pairs assigned smaller input-to-reference correction scores supervise earlier outputs, while pairs assigned larger scores supervise later outputs. Red points mark the assigned supervision endpoints. The final outputs show even better quality than their original labels with unstable quality.}
    \label{fig:label_quality_visualization}
\end{figure*}

UIE has progressed from handcrafted priors \cite{drews2016udcp,zhang2022mmle} to learning-based enhancement \cite{li2019uieb,peng2023ushape}, substantially improving restoration quality. Recent learning-based UIE methods improve enhancement through physics-guided modeling \cite{cong2023pugan,mei2025dpf}, contrastive objectives \cite{huang2023semiuir,liu2024ccl}, and specialized architectures \cite{peng2025ssuie,pucci2025cevae,xie2024uveb}, often relying on supervised learning where paired references specify which corrections a model should learn from degraded observations. Two widely used paired datasets are UIEB \cite{li2019uieb} and LSUI \cite{peng2023ushape}. Because matched images of the same underwater scene without the water medium are difficult to obtain, they construct references by evaluating and selecting candidate enhancements from multiple existing methods, a process involving substantial human effort. Human curation favors perceptually plausible results, yet the retained pairs demonstrate different degrees of correction relative to their inputs and may retain residual degradation (\cref{fig:label_quality_visualization}) even for the historical best result. Directly using these quality-unstable labels for end-to-end training may mislead the model about to what extent the degradations should be improved, resulting in sub-optimal performance. Restricting training to pairs with good quality is one possible response, but it removes examples that demonstrate modest corrections and reduces the available paired data, which can narrow the coverage of scene and degradation distribution under limited underwater paired training data. Nevertheless, expanding the curated collection requires further candidate generation and substantial human evaluation. These considerations motivate making fuller use of the corrections already represented in the existing pairs: We examine whether assigning references different supervision endpoints can improve the use of these heterogeneous corrections.

Our idea is to organize supervision according to the correction extent demonstrated by each input–reference pair: a reference that offers only partial correction can still provide useful supervision for an intermediate enhancement stage. A recurrent enhancer makes this possible by exposing a sequence of outputs, allowing references that demonstrate different degrees of correction to guide different stages of a shared enhancement process. Because the references have been selected by humans for perceptual plausibility, their changes relative to the inputs provide a practical cue for assigning these supervisory roles. We therefore use an allocation score to order the training pairs and quantize this ordering into supervision depths. Pairs with smaller estimated corrections supervise earlier outputs, while those with larger estimated corrections supervise later outputs. This assignment retains all original inputs and references, including pairs that demonstrate only modest corrections. We implement this principle through level-wise latent space enhancement (LLSE), which adapts a pretrained diffusion backbone into a shared recurrent latent enhancer. Each pair contributes a reference loss at its assigned endpoint, while all stages share enhancement parameters. At inference, every input traverses all $K$ stages. Finally, to compensate for local detail loss during VAE encoding and decoding, texture refinement via frequency modulation (TRFDM) combines the coarse output with texture cues retained in the raw image. In summary, our contributions are threefold:
\begin{enumerate}
    \item We propose RQUL-UIE, a supervision allocation framework that \textbf{r}evitalizes \textbf{q}uality-\textbf{u}nstable \textbf{l}abels by organizing the different corrections they demonstrate relative to their inputs. It retains all paired samples without requiring additional references or quality annotations.
    \item We design LLSE to allocate paired supervision across recurrent endpoints according to estimated input-to-reference correction extent. Each reference constrains one assigned output of a shared latent enhancer, retaining examples with both modest and larger corrections. TRFDM uses raw-image cues to compensate for detail loss caused by lossy compression.
    \item Evaluation on three datasets yields the best average performance as shown in \cref{fig:radar_overview}. Qualitative comparisons also show significantly improved visual quality.
\end{enumerate}


\section{Related Work}
\label{sec:literature}

\textbf{Underwater Image Formation Models.} Absorption and scattering govern underwater light propagation and produce wavelength-dependent attenuation and backscatter \cite{gonzalez2023survey,mcglamery1980computer,jaffe1990computer}. Classical formation models describe the observed image as attenuated scene radiance combined with veiling light \cite{chiang2011nerd}. Sea-thru \cite{akkaynak2018revised,akkaynak2019sea} further distinguishes the spectral and range dependence of attenuation and backscatter. These models provide physical foundations for understanding and correcting underwater degradations.
\\
\textbf{Underwater Image Enhancement Methods.} Handcrafted methods use Retinex or color-space priors to improve contrast and color consistency \cite{zhang2022mmle,zhuang2022laplacian}. Learning-based approaches instead recover enhanced images from data. Physics-guided methods estimate and decouple water effects using image formation models \cite{cong2023pugan,mei2025dpf}, while end-to-end methods develop architectures and training strategies for paired datasets. Semi-UIR \cite{huang2023semiuir} combines paired and unpaired samples through contrastive semi-supervised learning. HCLR \cite{zhou2024hclr} and CCL \cite{liu2024ccl} introduce contrastive regularization strategies, and WF-Diff \cite{zhao2024wfdiff} exploits frequency-domain processing. SS-UIE \cite{peng2025ssuie}, UVEB \cite{xie2024uveb}, and CE-VAE \cite{pucci2025cevae} further advance enhancement architectures. PF-USIE \cite{hu2025pfusie} constructs physics-guided synthetic underwater inputs paired with in-air targets and transfers enhancement to real images using a pretrained video diffusion model. SDAR-Net \cite{xu2026sdarnet} separates degradation styles from scene structure and uses adaptive soft routing for restoration.
\\
\textbf{Data Collection for Supervised Learning.} UIEB \cite{li2019uieb} and LSUI \cite{peng2023ushape} construct paired references through human selection among results from existing enhancement methods. This process combines the strengths of previous approaches and provides a prior of perceptual acceptability. The retained pairs demonstrate different degrees of correction relative to their inputs and can still contain residual degradation (\cref{fig:label_quality_visualization}). It motivates our study of how existing targets should be allocated during training. Synthetic datasets offer an alternative by applying underwater formation models to in-air images \cite{huang2023semiuir,hu2025pfusie}, although the synthetic-to-real gap can limit generalization \cite{hu2025pfusie}.
\\
\textbf{Pretrained Diffusion Models.} Diffusion models learn image distributions through denoising \cite{ho2020ddpm}, with latent diffusion reducing the spatial cost through VAE representations \cite{rombach2022sd}. These pretrained generative priors can be repurposed for image-conditioned prediction. Marigold \cite{ke2024repurposing} adapts Stable Diffusion for monocular depth estimation, while NormalCrafter \cite{bin2025normalcrafter} uses video diffusion priors for temporally consistent surface normal estimation. In underwater enhancement, PF-USIE \cite{hu2025pfusie} combines pretrained video diffusion priors with physics-based data synthesis to enhance image sequences. Garcia et al.\ \cite{garcia2025e2eft} further demonstrate deterministic task-specific adaptation through direct fine-tuning with reconstruction objectives. Together, these studies motivate reusing diffusion priors for supervised enhancement. We follow this route by adapting a pretrained latent diffusion backbone into a shared recurrent enhancer.

Existing studies improve enhancement models and reference construction, while diffusion-transfer studies show that pretrained image representations can support task-specific adaptation. These developments motivate examining the supervisory role assigned to each retained reference. RQUL-UIE organizes input-relative corrections across supervision endpoints of a shared recurrent latent enhancer and refines its coarse output using raw-image cues.

\section{Method}
\label{sec:method}
\begin{figure*}[htbp]
    \centering
    \includegraphics[width=\linewidth,pagebox=cropbox]{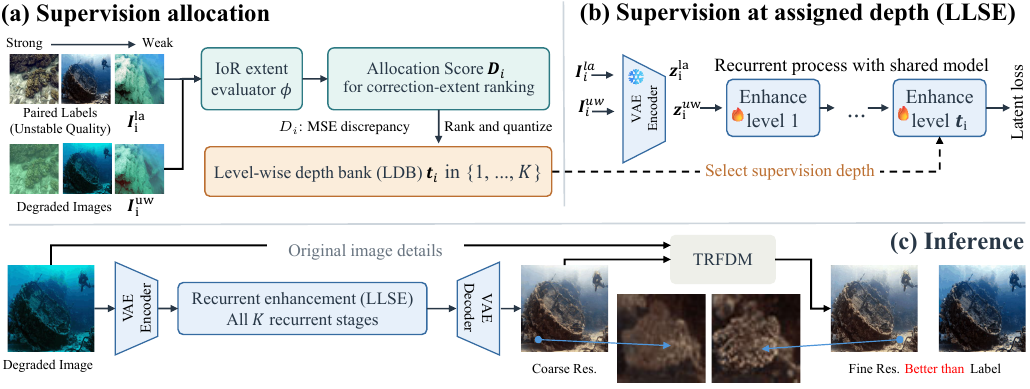}
    \caption{Overview of RQUL-UIE. An allocation score estimates the extent of input-to-reference (IoR) correction and assigns each training pair a supervision endpoint. Within LLSE, the reference constrains that output of a shared recurrent latent enhancer under the given supervision depth. At inference, every input traverses all $K$ enhancement levels, followed by TRFDM texture refinement using raw-image cues to get the final output that could be better than its reference with unstable quality.}
    \label{fig:framework}
\end{figure*}

RQUL-UIE organizes paired supervision by estimated input-to-reference correction extent (\cref{fig:framework}). An allocation score orders training pairs and assigns endpoints stored in the level-wise depth bank (LDB). Level-wise latent space enhancement (LLSE) supervises these endpoints with shared parameters across stages (\cref{sec:method_diff}). Inference traverses all $K$ levels, while texture refinement via frequency modulation (TRFDM) combines the decoded coarse output with raw-image cues to refine local details (\cref{sec:method_trfdm}).

\subsection{Preliminaries}\label{sec:pre_diff}


\noindent\textbf{Pretrained Latent Diffusion Backbone.}
Latent diffusion models \cite{rombach2022sd} combine a VAE \cite{kingma2013vae} with a conditional U-Net pretrained on large-scale image data. Our backbone is Stable Diffusion 2.1 (SD2.1)\footnote{https://huggingface.co/sd2-community/stable-diffusion-2-1}. Its frozen VAE maps images to latents $z=\mathcal E(I)$ and reconstructs them with decoder $\mathcal D$. Following the reuse of pretrained diffusion priors for depth estimation \cite{ke2024repurposing}, video normal estimation \cite{bin2025normalcrafter}, and underwater enhancement \cite{hu2025pfusie}, we initialize the enhancement U-Net from pretrained SD2.1 parameters. Its inherited time-conditioning interface distinguishes recurrent stages. 

We retain the backbone's time-conditioned U-Net $\mathcal U_\theta$ and $v$-prediction parameterization. Given a latent input $x_\tau$ and a pretrained time-conditioning index $\tau$, the corresponding clean-latent prediction is
\begin{equation}
    \hat z=\sqrt{\bar\alpha_\tau}\,x_\tau
    -\sqrt{1-\bar\alpha_\tau}\,v_\theta,
    \label{eq:latent_parameterization}
\end{equation}
where $\bar\alpha_\tau$ is the cumulative coefficient of the pretrained diffusion schedule. We adapt this parameterization for recurrent enhancement using a zero target-branch input and direct latent reconstruction supervision.

\subsection{Supervision Endpoint Allocation}\label{sec:method_diff}

References demonstrate different amounts of correction relative to their paired inputs. We use a practical estimate of this correction extent to allocate their supervision endpoints, assigning smaller scores to earlier recurrent outputs and larger scores to later outputs. 
Within LLSE, each pair contributes its reference loss at the assigned endpoint, and pairs assigned different depths jointly train the shared enhancer without any pairs discarded.

\begin{figure}[htbp]
    \centering
    \includegraphics[width=\linewidth]{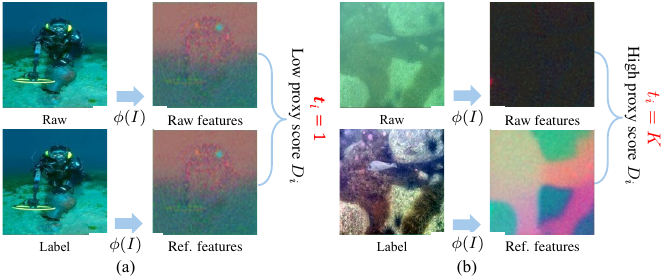}
    \caption{Examples of raw-reference pairs with low and high allocation scores. The default feature-discrepancy ranking determines the assigned supervision depth within the references. The features are visualized using the first three principal components.}
    \label{fig:rank_quant_vis}
\end{figure}

\noindent\textbf{Allocation Scoring and Depth Assignment.}
We instantiate the default allocation score using discrepancies between the final output tensors of the frozen pretrained SD2.1 U-Net due to its embedded semantic understanding. Because the references have already undergone human screening for basic perceptual plausibility, we use these output discrepancies as a practical proxy for the correction each pair demonstrates relative to its input. This choice reuses the same pretrained checkpoint that initializes our recurrent enhancer: the scoring representation remains frozen, while the enhancement U-Net is fine-tuned for UIE. Let $\phi$ denote this mapping from an image through VAE encoding to the U-Net output. For a raw underwater image $I_i^{uw}$ and its paired reference $I_i^{la}$, we compute
\begin{equation}
    D_i=\operatorname{MSE}\!\left(\phi(I_i^{uw}),\phi(I_i^{la})\right).
\end{equation}
The MSE averages squared differences over all output elements. The resulting scores define a relative allocation order within the curated collection and determine the assigned supervision depths:
\begin{equation}
    t_i=1+\left\lfloor
    \frac{r_i-1}{n}\,K
    \right\rfloor,
    \label{eq_ti}
\end{equation}
where $r_i\in\{1,\ldots,n\}$ is the one-based position of pair $i$ in the ascending score ordering, $n$ is the number of training pairs, and $K$ is both the maximum recurrent depth and the number of supervision levels. Each $t_i\in\{1,\ldots,K\}$ is the pair's assigned supervision endpoint. These assignments form the LDB, queried as $t_i=\mathrm Q_{LDB}(I_i^{uw})$. \cref{fig:rank_quant_vis} illustrates its two extreme levels based on the feature discrepancies.

\noindent\textbf{Diffusion-Pretrained Recurrent Latent Enhancement.}
Following the zero-noise input used in the deterministic task-specific adaptation of Garcia et al.\ \cite{garcia2025e2eft}, we set the target branch to zero. We apply this mapping recurrently: at each enhancement level, the conditioning branch receives the preceding output, starting from $z_i^{(0)}=\mathcal E(I_i^{uw})$. The recurrence index $k$ is mapped to the inherited pretrained time-conditioning index by
\begin{equation}
    \tau_k=\left\lfloor
    \frac{k}{K}(T-1)
    \right\rfloor,
\end{equation}
where $T$ is the number of pretrained diffusion time indices. For training sample $i$, the recurrence index runs over $k=1,\ldots,t_i$, with reference supervision applied at $k=t_i$. A single U-Net evaluation produces each enhancement level:
\begin{equation}
    \begin{aligned}
    v_i^{(k)}&=\mathcal U_\theta\!\left(
    [\operatorname{sg}(z_i^{(k-1)}),\mathbf 0],\tau_k,c_\varnothing\right),\\
    z_i^{(k)}&=-\sqrt{1-\bar\alpha_{\tau_k}}\,v_i^{(k)},
    \end{aligned}
    \label{eq:recurrent_enhancement}
\end{equation}
where $[\cdot,\cdot]$ denotes channel concatenation, $\mathbf 0$ has the same shape as the conditioning latent, $c_\varnothing$ is the empty-text embedding, and $\operatorname{sg}$ stops gradients through the preceding prediction. The second line follows from \cref{eq:latent_parameterization} with a zero target-branch input.

For sample $i$, we execute $t_i$ levels and apply the latent reconstruction term at its selected endpoint, using mean reduction over all latent elements:
\begin{equation}
    \mathcal L_{\mathrm{latent}}
    =\operatorname{MSE}\!\left(z_i^{(t_i)},\mathcal E(I_i^{la})\right).
    \label{eq:latent_supervision}
\end{equation}
The preceding levels generate the conditioning latent, and gradients are backpropagated through the selected supervision level. All levels share $\theta$. Within LLSE, each pair's reference loss is applied only at its assigned endpoint, making supervision depth a reference-dependent training choice. At inference, the network executes all $K$ levels, decodes $\mathcal D(z_i^{(K)})$ as the coarse image, and applies TRFDM using the raw image to recover texture details.

\begin{figure}[htbp]
    \centering
    \includegraphics[width=1\linewidth]{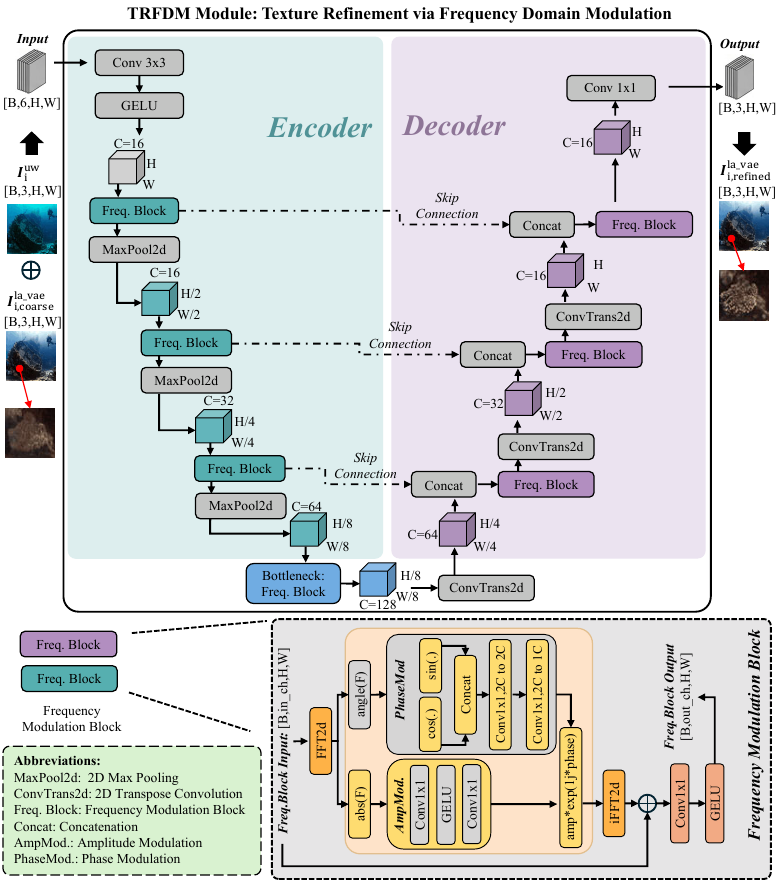}
    \caption{The proposed TRFDM module. A U-Net with frequency modulation blocks combines complementary image information through amplitude and phase modulation. The training input combines the raw image with its VAE-reconstructed reference; at inference, the latter is replaced by the decoded LLSE output.}
    \label{fig:fft_structure}
\end{figure}

\subsection{Frequency Modulation for Texture Refinement}\label{sec:method_trfdm}

\begin{figure}[htbp]
    \centering
    \includegraphics[width=1\linewidth]{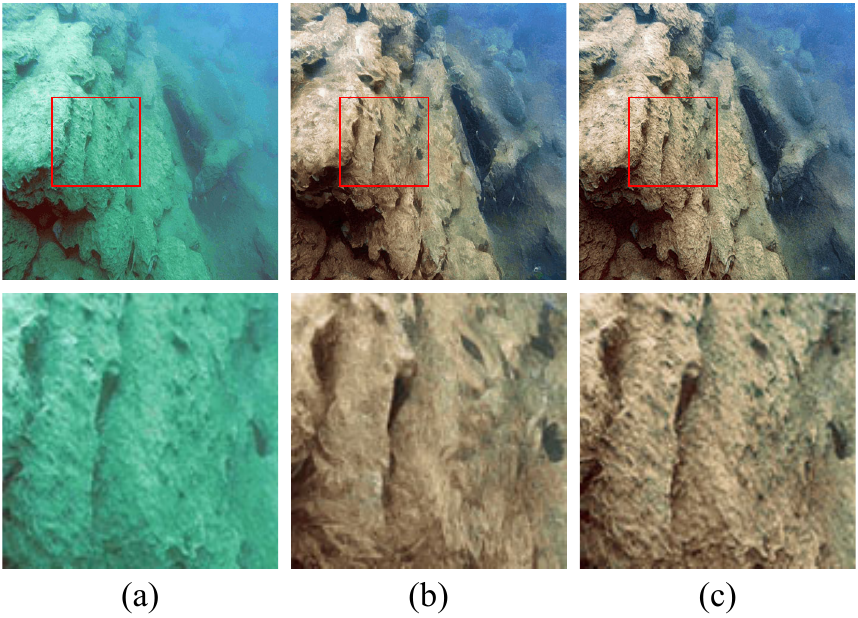}
    \caption{Texture refinement. The raw image (a) retains details smoothed in the coarse LLSE result (b). TRFDM combines these complementary representations to produce the refined output (c).}
    \label{fig:texture_from_raw}
\end{figure}
The decoded coarse enhancement can exhibit softened high-frequency details because VAE encoding and decoding involve lossy compression and reconstruction. Meanwhile, the raw underwater image retains complementary local textures despite its color and contrast degradations (\cref{fig:texture_from_raw}). TRFDM combines these textures with the global appearance of the coarse result through frequency-domain modulation. Its U-Net architecture uses a frequency-modulation block at each layer (\cref{fig:fft_structure}).

TRFDM is trained separately using a raw image and its VAE-reconstructed label $I_{i,coarse}^{la\_vae}=\mathcal D(\mathcal E(I_i^{la}))$. Their channel concatenation forms $X_{in}=I_i^{uw}\oplus I_{i,coarse}^{la\_vae}\in\mathbb R^{B\times6\times H\times W}$. For a feature map $X_{in}^{layer}$, the block computes its amplitude and phase as
\begin{align}
    A_X&=\operatorname{mag}(\operatorname{FFT}(X_{in}^{layer})),\\
    P_X&=\operatorname{angle}(\operatorname{FFT}(X_{in}^{layer})).
\end{align}
The amplitude is modulated by a Conv1$\times$1--GELU--Conv1$\times$1 structure. For phase modulation, concatenated $\cos(P_X)$ and $\sin(P_X)$ pass through two linear $1\times1$ convolutions. The reconstructed features pass through the U-Net to produce $I_{i,refined}^{la\_vae}=\operatorname{TRFDM}(X_{in})$. Training minimizes the pixel-space reconstruction loss
\begin{equation}
    \mathcal L_{freq}
    =\left\|I_i^{la}-I_{i,refined}^{la\_vae}\right\|_2^2.
\end{equation}

\begin{figure*}[htbp]
    \centering
    \includegraphics[width=\linewidth,height=0.92\textheight,keepaspectratio]{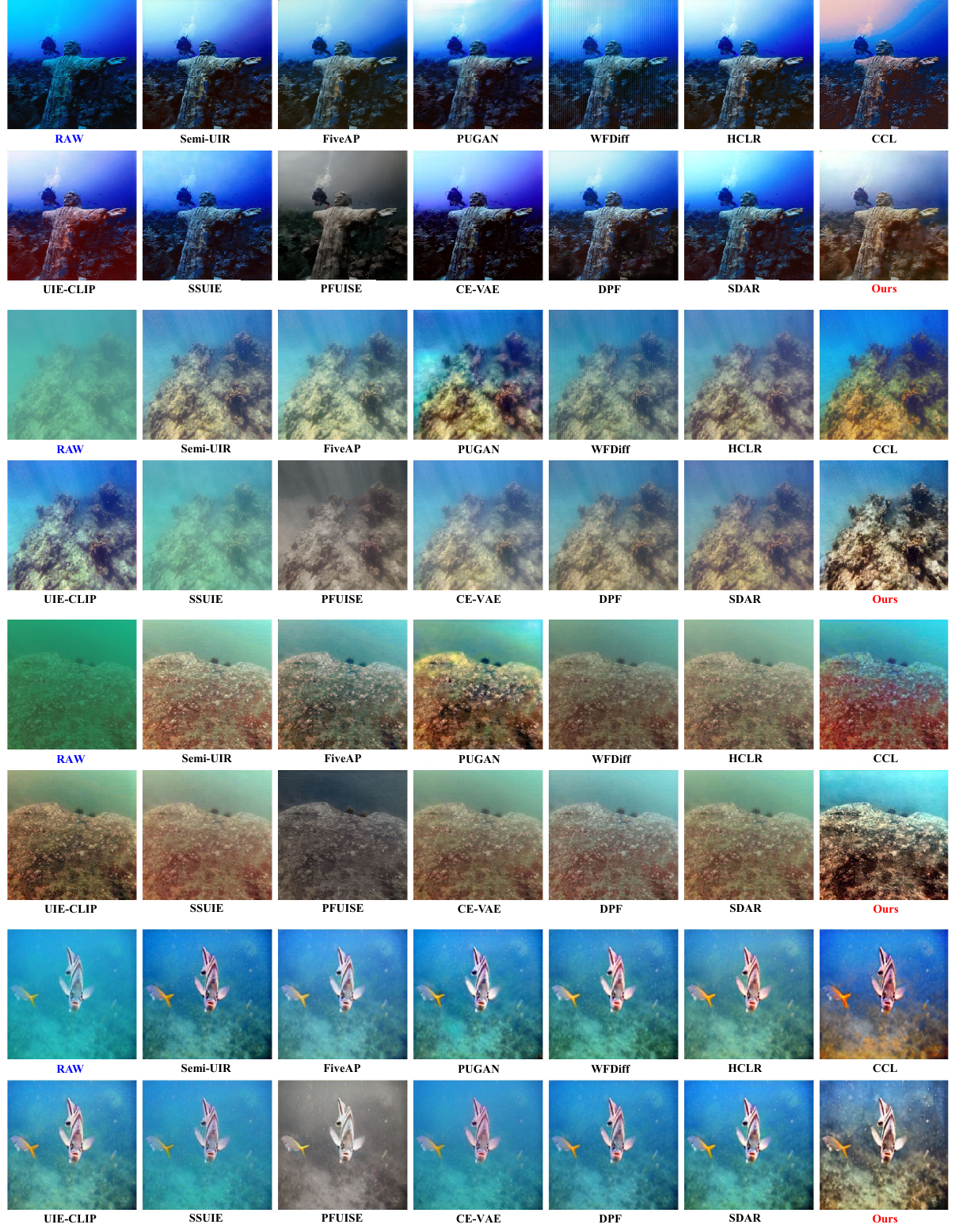}
    \caption{Visual comparison of UIE methods across four scene groups. ``Raw'' denotes the original underwater image.}
    \label{fig:vis_com_exp}
\end{figure*}

\begin{table*}[htbp]
\centering
\begingroup
\fontsize{7}{8}\selectfont
\setlength{\tabcolsep}{1pt}
\renewcommand{\arraystretch}{1.08}
\resizebox{\textwidth}{!}{
\begin{tabular}{@{}l|cccc|cccc|cccc|cc@{}}
\hlineB{2.5} 
\multirow{2}{*}{Method} & \multicolumn{4}{c|}{EUVP \cite{islam2020euvp}} & \multicolumn{4}{c|}{LSUI \cite{peng2023ushape}} & \multicolumn{4}{c|}{UIEB \cite{li2019uieb}} & \multicolumn{2}{c}{Avg. rank $\downarrow$}\\
\cline{2-15} 
 & MUSIQ $\uparrow$ & URANKER $\uparrow$ & \begin{tabular}[c]{@{}c@{}}\strut TWICE\\\strut MIX $\uparrow$\end{tabular} & UIQM $\uparrow$ & MUSIQ $\uparrow$ & URANKER $\uparrow$ & \begin{tabular}[c]{@{}c@{}}\strut TWICE\\\strut MIX $\uparrow$\end{tabular} & UIQM $\uparrow$ & MUSIQ $\uparrow$ & URANKER $\uparrow$ & \begin{tabular}[c]{@{}c@{}}\strut TWICE\\\strut MIX $\uparrow$\end{tabular} & UIQM $\uparrow$ & All & Cross \\
\hlineB{1.5} 
Semi-UIR \cite{huang2023semiuir} & \cellcolor{skyblue2}{50.410\,(02)} & \cellcolor{skyblue2}{2.472\,(02)} & 2.213\,(03) & 2.906\,(10.5) & \cellcolor{skyblue2}{43.268\,(02)} & \cellcolor{skyblue2}{2.365\,(02)} & 1.501\,(03) & 2.922\,(10) & 49.926\,(05) & 2.108\,(04) & 2.600\,(05) & 3.036\,(06) & \cellcolor{skyblue2}{4.54} & \cellcolor{skyblue2}{4.31} \\
FiveAP \cite{jiang2023fiveap}& 46.504\,(12) & 2.274\,(06) & 2.200\,(04) & 2.974\,(07) & 38.493\,(12) & 2.111\,(05) & 1.480\,(06) & 2.944\,(07) & 46.256\,(13) & 2.023\,(06) & 2.455\,(09) & 2.999\,(08.5) & 7.96 & 7.38 \\
PUGAN \cite{cong2023pugan}& 46.751\,(11) & 2.383\,(03) & 2.135\,(09) & \cellcolor{skyblue2}{3.147\,(02)} & 39.171\,(11) & 2.264\,(03) & \cellcolor{skyblue2}{1.508\,(02)} & 3.078\,(03) & 46.345\,(11) & \cellcolor{skyblue2}{2.200\,(02)} & 2.287\,(10) & 3.123\,(03) & 5.83 & 5.50 \\
HCLR \cite{zhou2024hclr} & 48.329\,(08) & 2.235\,(07) & 2.150\,(07) & 2.960\,(08) & 39.800\,(09) & 2.112\,(04) & 1.450\,(08) & 2.942\,(08) & 48.954\,(09) & 2.081\,(05) & 2.549\,(06) & 3.029\,(07) & 7.17 & 7.38 \\
WFDiff \cite{zhao2024wfdiff} & 46.111\,(13) & 2.057\,(10) & 1.831\,(12) & 2.933\,(09) & 38.434\,(13) & 1.805\,(12) & 1.166\,(13) & 2.920\,(11) & 47.700\,(10) & 1.777\,(13) & 2.012\,(12) & 2.931\,(12) & 11.67 & 11.63 \\
CCL \cite{liu2024ccl}& 48.638\,(06) & 2.051\,(11) & 2.146\,(08) & \cellcolor{skyblue1}{3.188\,(01)} & 40.662\,(06) & 1.954\,(10) & 1.412\,(10) & \cellcolor{skyblue1}{3.151\,(01)} & \cellcolor{skyblue1}{51.955\,(01)} & 1.859\,(11) & 2.644\,(04) & \cellcolor{skyblue1}{3.239\,(01)} & 5.83 & 6.63 \\
DPF \cite{mei2025dpf} & 47.976\,(09) & 2.196\,(08) & 2.124\,(10) & 3.083\,(05) & 40.314\,(08) & 2.040\,(07) & 1.420\,(09) & 3.020\,(05) & \cellcolor{skyblue2}{51.643\,(02)} & 1.905\,(09) & 2.498\,(07) & 3.108\,(05) & 7.00 & 7.63 \\
UIE-CLIP \cite{cao2025uieclip} & 48.747\,(05) & 2.164\,(09) & \cellcolor{skyblue2}{2.228\,(02)} & 3.108\,(03) & 40.565\,(07) & 2.012\,(08) & 1.494\,(05) & 3.037\,(04) & 49.578\,(06) & 2.017\,(07) & 2.645\,(03) & 3.120\,(04) & 5.25 & 5.38 \\
SSUIE \cite{peng2025ssuie} & 49.750\,(03) & 1.931\,(12) & 1.993\,(11) & 2.881\,(12) & 42.538\,(05) & 1.812\,(11) & 1.340\,(11) & 2.908\,(12) & 51.150\,(04) & 1.971\,(08) & 2.474\,(08) & 2.976\,(11) & 9.00 & 9.63 \\
PF-USIE \cite{hu2025pfusie}& 48.452\,(07) & 2.278\,(04.5) & 2.159\,(06) & 3.038\,(06) & 43.165\,(04) & 1.969\,(09) & 1.496\,(04) & 3.005\,(06) & 49.298\,(07) & 1.880\,(10) & 2.114\,(11) & 2.999\,(08.5) & 6.92 & 5.81 \\
CE-VAE \cite{pucci2025cevae}& 49.498\,(04) & 1.897\,(13) & 1.808\,(13) & 2.745\,(13) & \cellcolor{skyblue1}{43.919\,(01)} & 1.678\,(13) & 1.288\,(12) & 2.941\,(09) & 46.304\,(12) & 1.849\,(12) & 1.879\,(13) & 2.877\,(13) & 10.67 & 9.75 \\
SDAR \cite{xu2026sdarnet}& 47.970\,(10) & 2.278\,(04.5) & 2.178\,(05) & 2.906\,(10.5) & 39.549\,(10) & 2.095\,(06) & 1.468\,(07) & 2.885\,(13) & 49.102\,(08) & 2.131\,(03) & \cellcolor{skyblue1}{2.677\,(01)} & 2.985\,(10) & 7.33 & 8.25 \\
\hlineB{1.5}
\textbf{RQUL} & \cellcolor{skyblue1}{50.500\,(01)} & \cellcolor{skyblue1}{2.802\,(01)} & \cellcolor{skyblue1}{2.337\,(01)} & 3.096\,(04) & 43.223\,(03) & \cellcolor{skyblue1}{2.808\,(01)} & \cellcolor{skyblue1}{1.723\,(01)} & \cellcolor{skyblue2}{3.118\,(02)} & 51.634\,(03) & \cellcolor{skyblue1}{2.408\,(01)} & \cellcolor{skyblue2}{2.668\,(02)} & \cellcolor{skyblue2}{3.148\,(02)} & \cellcolor{skyblue1}{1.83} & \cellcolor{skyblue1}{1.75} \\
\end{tabular}
}
\endgroup
\caption{Quantitative comparison. RQUL-UIE trains on complete UIEB; its UIEB results assess source training inputs, while EUVP/LSUI provide cross-dataset comparisons. All averages twelve dataset--metric ranks; Cross averages the eight EUVP/LSUI ranks for every method. Parentheses give metric ranks. The \colorbox{skyblue1}{best} and \colorbox{skyblue2}{second-best} values are highlighted per column.}
\label{tab:main_comparison}
\end{table*}

\raggedbottom
\section{Experiments}
\label{sec:exp}
\subsection{Experiment Setup}
We implement RQUL-UIE in PyTorch and use the complete UIEB \cite{li2019uieb} collection as the default training source for both LLSE and TRFDM modules. LLSE adapts the pretrained SD2.1 U-Net with $K=9$ supervision levels and a maximum recurrence depth of nine. Each training pair supervises the endpoint selected by its assigned supervision depth $t_i$. Inference executes all nine levels, with one U-Net evaluation per level. We train the enhancement backbone on two RTX 3090 GPUs for 80k iterations (batch size=1). TRFDM is trained separately for 4k epochs (batch size=16). The learning rates are $3\times10^{-5}$ and $2\times10^{-4}$ for the enhancement backbone and TRFDM, respectively, with optimizer coefficients $\beta_1=0.9$ and $\beta_2=0.999$. The evaluation resolution is fixed to $256\times256$.

\subsection{Evaluation}

\textbf{Evaluation Protocols.} The model is evaluated on UIEB \cite{li2019uieb}, EUVP \cite{islam2020euvp}, and LSUI \cite{peng2023ushape}. UIEB results assess the source training collection, while EUVP and LSUI provide cross-dataset comparisons. We compare with 12 methods. All four evaluation metrics---MUSIQ \cite{ke2021musiq}, URANKER \cite{guo2023uranker}, TWICEMIX \cite{fu2022twicemix}, and UIQM \cite{panetta2015uiqm}---are higher-is-better. We summarize the results using both all-collection and cross-dataset average ranks:
\begin{equation}
\begin{aligned}
R_{\mathrm{all}}(a)&=\frac{1}{12}\sum_{d\in\mathcal D_{\mathrm{all}}}\sum_{m=1}^{4}r_{a,d,m},\\
R_{\mathrm{cross}}(a;s)&=\frac{1}{8}\sum_{d\in\mathcal D_{\mathrm{cross}}(s)}\sum_{m=1}^{4}r_{a,d,m}.
\end{aligned}
\label{eq:average_ranks}
\end{equation}
Here, $a$ denotes a compared method or ablation configuration, $d$ denotes an evaluation dataset, and $m\in\{1,2,3,4\}$ indexes the above four metrics. For each dataset--metric pair $(d,m)$, $r_{a,d,m}$ is the rank of configuration $a$ among the rows in the current comparison table, with rank 1 assigned to the highest score and average ranks used for ties. The variable $s$ denotes the training-source dataset of our model for that comparison. We define $\mathcal D_{\mathrm{all}}=\{\mathrm{EUVP},\mathrm{LSUI},\mathrm{UIEB}\}$ and $\mathcal D_{\mathrm{cross}}(s)=\mathcal D_{\mathrm{all}}\setminus\{s\}$. For the UIEB-trained configurations evaluated here, $s=\mathrm{UIEB}$ and $\mathcal D_{\mathrm{cross}}(s)=\{\mathrm{EUVP},\mathrm{LSUI}\}$. Every compared row, including external baselines, uses the same aggregation columns. Lower average ranks indicate better performance.

\noindent\textbf{Quantitative Comparison.} \cref{tab:main_comparison} compares the trained model with 12 enhancement methods on EUVP, LSUI, and UIEB. On EUVP and LSUI, RQUL-UIE achieves the best cross-dataset average rank of 1.75, compared with 4.31 for Semi-UIR \cite{huang2023semiuir}, and ranks among the top two in six of the eight dataset--metric combinations. Including the UIEB source collection, it also achieves the best All average rank of 1.83 versus 4.54. These results show a clear aggregate advantage in both the cross-dataset evaluation and the full reported comparison and suppport the effectiveness of the proposed supervision allocation strategy.


\begin{table*}[!t]
\centering
\begingroup
\fontsize{7}{8}\selectfont
\setlength{\tabcolsep}{1pt}
\renewcommand{\arraystretch}{1.08}
\resizebox{\textwidth}{!}{
\begin{tabular}{@{}l|cccc|cccc|cccc|cc@{}}
\hlineB{2.5}
\multirow{2}{*}{Configuration} & \multicolumn{4}{c|}{EUVP \cite{islam2020euvp}} & \multicolumn{4}{c|}{LSUI \cite{peng2023ushape}} & \multicolumn{4}{c|}{UIEB \cite{li2019uieb}} & \multicolumn{2}{c}{Avg. rank $\downarrow$}\\
\cline{2-15}
 & MUSIQ $\uparrow$ & URANKER $\uparrow$ & TWICEMIX $\uparrow$ & UIQM $\uparrow$ & MUSIQ $\uparrow$ & URANKER $\uparrow$ & TWICEMIX $\uparrow$ & UIQM $\uparrow$ & MUSIQ $\uparrow$ & URANKER $\uparrow$ & TWICEMIX $\uparrow$ & UIQM $\uparrow$ & All & Cross \\
\hlineB{1.5}
Semi-UIR \cite{huang2023semiuir} & \cellcolor{skyblue2}{50.410\,(2)} & 2.472\,(7) & 2.213\,(6) & 2.906\,(4) & \cellcolor{skyblue1}{43.268\,(1)} & 2.365\,(6) & 1.501\,(8) & 2.922\,(4) & 49.926\,(5) & 2.108\,(8) & 2.600\,(5) & 3.036\,(4) & 5.00 & 4.75 \\
Best-subset & 49.890\,(3) & 2.553\,(4) & 2.309\,(3) & 3.022\,(3) & 41.570\,(4) & 2.403\,(5) & 1.657\,(3) & 3.023\,(3) & \cellcolor{skyblue2}{51.443\,(2)} & \cellcolor{skyblue2}{2.345\,(2)} & \cellcolor{skyblue1}{2.753\,(1)} & 3.113\,(3) & 3.00 & 3.50 \\
W/o TRFDM & 47.269\,(8) & \cellcolor{skyblue2}{2.775\,(2)} & \cellcolor{skyblue1}{2.525\,(1)} & \cellcolor{skyblue1}{3.316\,(1)} & 42.326\,(3) & \cellcolor{skyblue2}{2.789\,(2)} & \cellcolor{skyblue1}{1.961\,(1)} & \cellcolor{skyblue1}{3.298\,(1)} & 47.236\,(8) & 2.322\,(3) & \cellcolor{skyblue2}{2.672\,(2)} & \cellcolor{skyblue1}{3.249\,(1)} & \cellcolor{skyblue2}{2.75} & \cellcolor{skyblue2}{2.38} \\
\begin{tabular}[c]{@{}l@{}}W/o LLSE (equal allocation)\end{tabular} & 48.462\,(7) & 2.422\,(8) & 2.256\,(4) & 2.859\,(6.5) & 40.357\,(8) & 2.258\,(8) & 1.522\,(7) & 2.921\,(5) & 50.692\,(4) & 2.165\,(5) & 2.664\,(4) & 3.005\,(6) & 6.04 & 6.69 \\
Random allocation & 48.487\,(6) & 2.517\,(6) & 2.126\,(8) & 2.864\,(5) & 41.153\,(6) & 2.482\,(4) & 1.587\,(5) & 2.794\,(8) & 49.511\,(7) & 2.121\,(7) & 2.353\,(8) & 2.668\,(8) & 6.50 & 6.00 \\
Reversed allocation & 49.372\,(4) & 2.604\,(3) & 2.219\,(5) & 2.856\,(8) & 41.508\,(5) & 2.631\,(3) & 1.617\,(4) & 2.842\,(6) & 50.863\,(3) & 2.205\,(4) & 2.588\,(6) & 3.021\,(5) & 4.67 & 4.75 \\
\begin{tabular}[c]{@{}l@{}}Weighted equal allocation\end{tabular} & 48.874\,(5) & 2.526\,(5) & 2.153\,(7) & 2.859\,(6.5) & 40.726\,(7) & 2.358\,(7) & 1.541\,(6) & 2.807\,(7) & 49.692\,(6) & 2.145\,(6) & 2.543\,(7) & 2.818\,(7) & 6.38 & 6.31 \\
\hlineB{2.5}
Full model & \cellcolor{skyblue1}{50.500\,(1)} & \cellcolor{skyblue1}{2.802\,(1)} & \cellcolor{skyblue2}{2.337\,(2)} & \cellcolor{skyblue2}{3.096\,(2)} & \cellcolor{skyblue2}{43.223\,(2)} & \cellcolor{skyblue1}{2.808\,(1)} & \cellcolor{skyblue2}{1.723\,(2)} & \cellcolor{skyblue2}{3.118\,(2)} & \cellcolor{skyblue1}{51.634\,(1)} & \cellcolor{skyblue1}{2.408\,(1)} & 2.668\,(3) & \cellcolor{skyblue2}{3.148\,(2)} & \cellcolor{skyblue1}{1.67} & \cellcolor{skyblue1}{1.63} \\

\end{tabular}
}
\endgroup
\caption{Ablations of allocation, sample selection, and refinement. UIEB is the source collection for the proposed configurations, trained on all pairs or the specified subset. Semi-UIR is an external baseline for comparison. All averages twelve dataset--metric ranks; Cross averages the eight EUVP/LSUI ranks for every row. All eight rows are ranked jointly; parentheses give metric ranks. The \colorbox{skyblue1}{best} and \colorbox{skyblue2}{second-best} values are highlighted per column.}
\label{tab:supervision_allocation}
\end{table*}

\begin{figure}[!t]
    \centering
    \includegraphics[width=\linewidth]{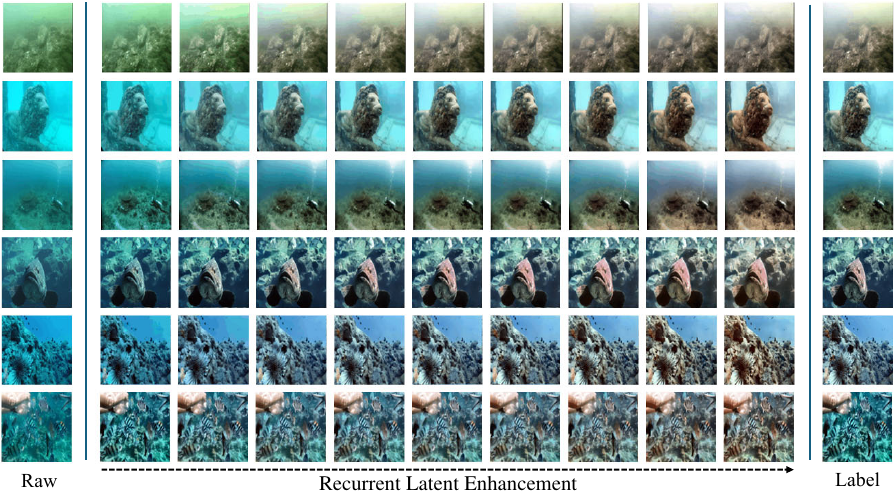}
    \caption{Recurrent enhancement for six selected inputs (columns). Rows show raw images, outputs at different stages ($K=9$), and paired references.}
    \label{fig:vis_mddp_exp}
\end{figure}

\noindent\textbf{Qualitative Comparison.} As shown in \cref{fig:vis_com_exp}, RQUL-UIE reduces global color casts while improving the visibility of structures obscured by underwater degradation. These improvements are particularly evident in the second and fourth scene groups: blue or green background biases are reduced, and distant textures that are difficult to distinguish in the raw images become more clearly visible. Compared with the other displayed methods, our results more effectively combine color correction with improved visibility in hazy, low-contrast regions. Several competing results retain noticeable color bias or leave background structures weakly defined, illustrating the difficulty of addressing these degradations simultaneously in the displayed scenes.

\cref{fig:vis_mddp_exp} presents recurrent enhancement of selected underwater inputs alongside their paired references. In these examples, the final outputs exhibit more correct colors and clearer structures than the corresponding references, illustrating how the learned enhancer can make further use of the corrections represented in the training data. These observations support assigning references with different estimated correction extents to different supervision endpoints of the shared enhancer. In addition, TRFDM further complements the coarse output with local information retained in the raw image for texture refinement. In the enlarged regions of \cref{fig:texturefromraw_ab}, the refined result shows more correct and clearer textures than the coarse enhancement.


\subsection{Ablation Study}

\noindent\textbf{Supervision Allocation Ablation.}
\cref{tab:supervision_allocation} tests how each reference should supervise the recurrent enhancer. Equal allocation uses a common final endpoint; random allocation is balanced across levels, assigned once per image, and fixed throughout training; reversed allocation inverts the score-derived ordering. Weighted equal allocation uses $t_i/K$ to scale the paired loss at the common endpoint. These controls retain all pairs, TRFDM, and the SD2.1 enhancement backbone. The full model exceeds all four controls across the eight cross-dataset metrics. The random and reversed comparisons support using pair-derived information and its ordering, while the weighted comparison favors selecting supervision endpoints over the tested loss-weighting rule.

\begin{figure}[htbp]
    \centering
    \includegraphics[width=1\linewidth]{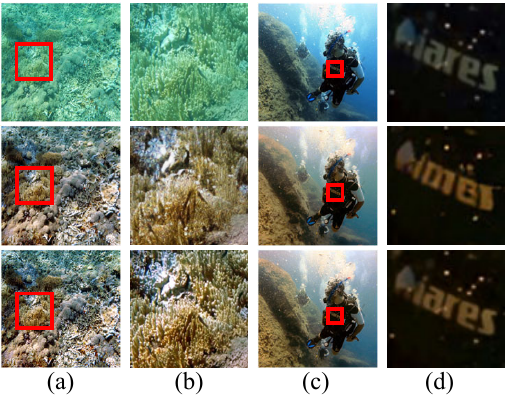}
    \caption{Texture refinement ablation. First row: degraded images; second row: coarse enhancement; third row: refined output. Columns (a), (c): two scenes; columns (b), (d): enlarged views of the corresponding red regions.}
    \label{fig:texturefromraw_ab}
\end{figure}

\noindent\textbf{Selection and Refinement Ablation.}
Best-subset retains the highest allocation-score bin, which means $t_i=9$. Retaining all pairs with the proposed allocation achieves better average ranks than training on this selected subset, supporting fuller use of the available paired supervision. With allocation fixed, TRFDM strengthens the aggregate performance. The local textures that are clearer and more consistent with the raw image in \cref{fig:texturefromraw_ab} further illustrate how raw-image cues complement the coarse enhancement.

\begin{samepage}
\section{Conclusion}
\label{sec:conclusion}
We presented RQUL-UIE, a framework that reallocates existing paired references that have unstable enhancement quality to recurrent supervision endpoints according to estimated input-relative correction extent. It allows a shared enhancer to learn from different degrees of correction while retaining all pairs and avoiding additional manual reference assessment. In this way, references demonstrating limited corrections can still provide useful supervision at intermediate enhancement stages. TRFDM further improves local detail using raw-image texture cues. Quantitative comparisons demonstrate strong overall performance across the evaluated datasets, while visual results show improved color balance and clearer structures. 

\noindent\textbf{Limitations.} The current recurrent framework incurs substantial inference latency, making it unsuitable for real-time deployment. Future work will focus on accelerating inference while preserving enhancement quality.

\par\end{samepage}

\clearpage
\twocolumn
\raggedbottom
\section*{LLM Usage Statement}
OpenAI Codex assisted with drafting and revising the manuscript and supplementary material, organizing methodological and experimental explanations, checking numerical consistency and rank calculations from author-provided measurements, and preparing framework diagrams, figure annotations, and LaTeX formatting. The authors are responsible for the research design, experimental data, analyses, scientific claims, and conclusions.

{
    \small
    \bibliographystyle{ieeenat_fullname}
    \bibliography{main}

@String(ICCV= {Int. Conf. Comput. Vis.})

@String(AAAI = {AAAI})

@String(ICCV  = {ICCV})

@article{gonzalez2023survey,
  title={A survey on underwater computer vision},
  author={Gonz{\'a}lez-Sabbagh, Salma P and Robles-Kelly, Antonio},
  journal={ACM Computing Surveys},
  volume={55},
  number={13s},
  pages={1--39},
  year={2023},
}

@inproceedings{mcglamery1980computer,
  title={A computer model for underwater camera systems},
  author={McGlamery, BL},
  booktitle={Ocean Optics VI},
  volume={208},
  pages={221--231},
  year={1980},
  organization={SPIE}
}

@article{jaffe1990computer,
  title={Computer modeling and the design of optimal underwater imaging systems},
  author={Jaffe, Jules S},
  journal={IEEE Journal of Oceanic Engineering},
  volume={15},
  number={2},
  pages={101--111},
  year={1990},
  publisher={IEEE},
}

@inproceedings{akkaynak2018revised,
  title={A revised underwater image formation model},
  author={Akkaynak, Derya and Treibitz, Tali},
  booktitle={Proceedings of the IEEE conference on computer vision and pattern recognition},
  pages={6723--6732},
  year={2018}
}

@article{drews2016udcp,
  title={Underwater depth estimation and image restoration based on single images},
  author={Drews, Paulo LJ and Nascimento, Erickson R and Botelho, Silvia SC and Campos, Mario Fernando Montenegro},
  journal={IEEE computer graphics and applications},
  volume={36},
  number={2},
  pages={24--35},
  year={2016},
  publisher={IEEE}
}

@inproceedings{akkaynak2019sea,
  title={Sea-thru: A method for removing water from underwater images},
  author={Akkaynak, Derya and Treibitz, Tali},
  booktitle={Proceedings of the IEEE/CVF conference on computer vision and pattern recognition},
  pages={1682--1691},
  year={2019}
}

@article{zhuang2022laplacian,
  title={Underwater image enhancement with hyper-laplacian reflectance priors},
  author={Zhuang, Peixian and Wu, Jiamin and Porikli, Fatih and Li, Chongyi},
  journal={IEEE Transactions on Image Processing},
  volume={31},
  pages={5442--5455},
  year={2022},
  publisher={IEEE}
}

@article{zhang2022mmle,
  title={Underwater image enhancement via minimal color loss and locally adaptive contrast enhancement},
  author={Zhang, Weidong and Zhuang, Peixian and Sun, Hai-Han and Li, Guohou and Kwong, Sam and Li, Chongyi},
  journal={IEEE Transactions on Image Processing},
  volume={31},
  pages={3997--4010},
  year={2022},
  publisher={IEEE}
}

@article{cong2023pugan,
  title={Pugan: Physical model-guided underwater image enhancement using gan with dual-discriminators},
  author={Cong, Runmin and Yang, Wenyu and Zhang, Wei and Li, Chongyi and Guo, Chun-Le and Huang, Qingming and Kwong, Sam},
  journal={IEEE Transactions on Image Processing},
  volume={32},
  pages={4472--4485},
  year={2023},
  publisher={IEEE}
}

@inproceedings{huang2023semiuir,
  title={Contrastive semi-supervised learning for underwater image restoration via reliable bank},
  author={Huang, Shirui and Wang, Keyan and Liu, Huan and Chen, Jun and Li, Yunsong},
  booktitle={Proceedings of the IEEE/CVF conference on computer vision and pattern recognition},
  pages={18145--18155},
  year={2023}
}

@article{jiang2023fiveap,
  title={Five A+ Network: You Only Need 9K Parameters for Underwater Image Enhancement},
  author={Jiang, Jingxia and Ye, Tian and Bai, Jinbin and Chen, Sixiang and Chai, Wenhao and Jun, Shi and Liu, Yun and Chen, Erkang},
  journal={arXiv preprint arXiv:2305.08824},
  year={2023}
}

@inproceedings{xie2024uveb,
  title={UVEB: A Large-scale Benchmark and Baseline Towards Real-World Underwater Video Enhancement},
  author={Xie, Yaofeng and Kong, Lingwei and Chen, Kai and Zheng, Ziqiang and Yu, Xiao and Yu, Zhibin and Zheng, Bing},
  booktitle={Proceedings of the IEEE/CVF Conference on Computer Vision and Pattern Recognition},
  pages={22358--22367},
  year={2024}
}

@article{islam2020euvp,
  title={Fast underwater image enhancement for improved visual perception},
  author={Islam, Md Jahidul and Xia, Youya and Sattar, Junaed},
  journal={IEEE Robotics and Automation Letters},
  volume={5},
  number={2},
  pages={3227--3234},
  year={2020},
  publisher={IEEE}
}

@article{li2019uieb,
  title={An underwater image enhancement benchmark dataset and beyond},
  author={Li, Chongyi and Guo, Chunle and Ren, Wenqi and Cong, Runmin and Hou, Junhui and Kwong, Sam and Tao, Dacheng},
  journal={IEEE transactions on image processing},
  volume={29},
  pages={4376--4389},
  year={2019},
  publisher={IEEE}
}

@inproceedings{rombach2022sd,
  title={High-resolution image synthesis with latent diffusion models},
  author={Rombach, Robin and Blattmann, Andreas and Lorenz, Dominik and Esser, Patrick and Ommer, Bj{\"o}rn},
  booktitle={Proceedings of the IEEE/CVF conference on computer vision and pattern recognition},
  pages={10684--10695},
  year={2022}
}

@article{ho2020ddpm,
  title={Denoising diffusion probabilistic models},
  author={Ho, Jonathan and Jain, Ajay and Abbeel, Pieter},
  journal={Advances in neural information processing systems},
  volume={33},
  pages={6840--6851},
  year={2020}
}

@article{chiang2011nerd,
  title={Underwater image enhancement by wavelength compensation and dehazing},
  author={Chiang, John Y and Chen, Ying-Ching},
  journal={IEEE transactions on image processing},
  volume={21},
  number={4},
  pages={1756--1769},
  year={2011},
  publisher={IEEE}
}

@article{kingma2013vae,
  title={Auto-encoding variational bayes},
  author={Kingma, Diederik P},
  journal={arXiv preprint arXiv:1312.6114},
  year={2013}
}

@article{panetta2015uiqm,
  title={Human-visual-system-inspired underwater image quality measures},
  author={Panetta, Karen and Gao, Chen and Agaian, Sos},
  journal={IEEE Journal of Oceanic Engineering},
  volume={41},
  number={3},
  pages={541--551},
  year={2015},
  publisher={IEEE}
}

@inproceedings{ke2021musiq,
  title={Musiq: Multi-scale image quality transformer},
  author={Ke, Junjie and Wang, Qifei and Wang, Yilin and Milanfar, Peyman and Yang, Feng},
  booktitle={Proceedings of the IEEE/CVF international conference on computer vision},
  pages={5148--5157},
  year={2021}
}

@inproceedings{guo2023uranker,
  title={Underwater ranker: Learn which is better and how to be better},
  author={Guo, Chunle and Wu, Ruiqi and Jin, Xin and Han, Linghao and Zhang, Weidong and Chai, Zhi and Li, Chongyi},
  booktitle={Proceedings of the AAAI conference on artificial intelligence},
  volume={37},
  number={1},
  pages={702--709},
  year={2023}
}

@inproceedings{ke2024repurposing,
  title={Repurposing diffusion-based image generators for monocular depth estimation},
  author={Ke, Bingxin and Obukhov, Anton and Huang, Shengyu and Metzger, Nando and Daudt, Rodrigo Caye and Schindler, Konrad},
  booktitle={Proceedings of the IEEE/CVF Conference on Computer Vision and Pattern Recognition},
  pages={9492--9502},
  year={2024}
}

@article{liu2024ccl,
  title={Underwater Image Enhancement with Cascaded Contrastive Learning},
  author={Liu, Yi and Jiang, Qiuping and Wang, Xinyi and Luo, Ting and Zhou, Jingchun},
  journal={IEEE Transactions on Multimedia},
  year={2024},
  publisher={IEEE}
}

@article{zhou2024hclr,
  title={HCLR-Net: hybrid contrastive learning regularization with locally randomized perturbation for underwater image enhancement},
  author={Zhou, Jingchun and Sun, Jiaming and Li, Chongyi and Jiang, Qiuping and Zhou, Man and Lam, Kin-Man and Zhang, Weishi and Fu, Xianping},
  journal={International Journal of Computer Vision},
  volume={132},
  number={10},
  pages={4132--4156},
  year={2024},
  publisher={Springer}
}

@inproceedings{peng2025ssuie,
  title={Adaptive Dual-domain Learning for Underwater Image Enhancement},
  author={Peng, Lintao and Bian, Liheng},
  booktitle={Proceedings of the AAAI Conference on Artificial Intelligence},
  volume={39},
  number={6},
  pages={6461--6469},
  year={2025}
}

@inproceedings{pucci2025cevae,
  title={Ce-vae: Capsule enhanced variational autoencoder for underwater image enhancement},
  author={Pucci, Rita and Martinel, Niki},
  booktitle={2025 IEEE/CVF Winter Conference on Applications of Computer Vision (WACV)},
  pages={2113--2123},
  year={2025},
  organization={IEEE}
}

@article{mei2025dpf,
  title={DPF-Net: Physical Imaging Model Embedded Data-Driven Underwater Image Enhancement},
  author={Mei, Han and Li, Kunqian and Liu, Shuaixin and Ma, Chengzhi and Jiang, Qianli},
  journal={ISPRS Journal of Photogrammetry and Remote Sensing},
  volume={228},
  pages={679-693},
  year={2025}
}

@inproceedings{ravi2025sam2,
  title={{SAM 2}: Segment Anything in Images and Videos},
  author={Ravi, Nikhila and Gabeur, Valentin and Hu, Yuan-Ting and Hu, Ronghang and Ryali, Chaitanya and Ma, Tengyu and Khedr, Haitham and R{\"a}dle, Roman and Rolland, Chloe and Gustafson, Laura and others},
  booktitle={International Conference on Learning Representations},
  volume={2025},
  year={2025}
}

@article{oquab2023dinov2,
  title={{DINOv2}: Learning Robust Visual Features without Supervision},
  author={Oquab, Maxime and Darcet, Timoth{\'e}e and Moutakanni, Th{\'e}o and Vo, Huy V. and Szafraniec, Marc and Khalidov, Vasil and Fernandez, Pierre and Haziza, Daniel and Massa, Francisco and El-Nouby, Alaaeldin and others},
  journal={arXiv preprint arXiv:2304.07193},
  year={2023}
}

@article{peng2023ushape,
  title={U-shape transformer for underwater image enhancement},
  author={Peng, Lintao and Zhu, Chunli and Bian, Liheng},
  journal={IEEE Transactions on Image Processing},
  volume={32},
  pages={3066--3079},
  year={2023},
  publisher={IEEE}
}

@inproceedings{zhao2024wfdiff,
  title={Wavelet-based fourier information interaction with frequency diffusion adjustment for underwater image restoration},
  author={Zhao, Chen and Cai, Weiling and Dong, Chenyu and Hu, Chengwei},
  booktitle={Proceedings of the IEEE/CVF Conference on Computer Vision and Pattern Recognition},
  pages={8281--8291},
  year={2024}
}

@inproceedings{garcia2025e2eft,
  title={Fine-tuning image-conditional diffusion models is easier than you think},
  author={Garcia, Gonzalo Martin and Abou Zeid, Karim and Schmidt, Christian and De Geus, Daan and Hermans, Alexander and Leibe, Bastian},
  booktitle={2025 IEEE/CVF Winter Conference on Applications of Computer Vision (WACV)},
  pages={753--762},
  year={2025},
  organization={IEEE}
}

@article{cao2025uieclip,
  title={Unveiling the underwater world: CLIP perception model-guided underwater image enhancement},
  author={Cao, Jiangzhong and Zeng, Zekai and Zhang, Xu and Zhang, Huan and Fan, Chunling and Jiang, Gangyi and Lin, Weisi},
  journal={Pattern Recognition},
  volume={162},
  pages={111395},
  year={2025},
  publisher={Elsevier}
}

@article{hu2025pfusie,
  title={Underwater sequential images enhancement via diffusion and physics priors fusion},
  author={Hu, Haochen and Bin, Yanrui and Wen, Chih-yung and Wang, Bing},
  journal={Information Fusion},
  pages={103365},
  year={2025},
  publisher={Elsevier}
}

@article{xu2026sdarnet,
  title={Style-Decoupled Adaptive Routing Network for Underwater Image Enhancement},
  author={Xu, Hang and Long, Chen and Wang, Bing and Chen, Hao and Dong, Zhen},
  journal={arXiv preprint arXiv:2604.12257},
  year={2026}
}

@article{fu2022twicemix,
  title={Twice mixing: A rank learning based quality assessment approach for underwater image enhancement},
  author={Fu, Zhenqi and Fu, Xueyang and Huang, Yue and Ding, Xinghao},
  journal={Signal Processing: Image Communication},
  volume={102},
  pages={116622},
  year={2022},
  publisher={Elsevier}
}

@inproceedings{bin2025normalcrafter,
  author={Bin, Yanrui and Hu, Wenbo and Wang, Haoyuan and Chen, Xinya and Wang, Bing},
  title={{NormalCrafter}: Learning Temporally Consistent Normals from Video Diffusion Priors},
  booktitle={Proceedings of the IEEE/CVF International Conference on Computer Vision (ICCV)},
  month={October},
  year={2025},
  pages={8330--8339},
  url={https://openaccess.thecvf.com/content/ICCV2025/html/Bin_NormalCrafter_Learning_Temporally_Consistent_Normals_from_Video_Diffusion_Priors_ICCV_2025_paper.html}
}
}

\clearpage
\maketitlesupplementary
\label{sec:supplementary_start}

\section{Additional Ablation Studies}

\subsection{Supervision-Allocation Protocols}
\label{sec:sup_assignment_protocols}

\noindent\textbf{Weighted equal allocation.}
This control uses the same allocation-score-derived levels as the full model, applying them as sample weights in equal allocation at a common final endpoint. For sample $i$, the weight and weighted paired-target term are
\begin{equation}
    w_i=\frac{t_i}{K},\qquad
    \ell_i^{\mathrm{weighted}}=w_i\,\ell_i^{\mathrm{paired}},
    \label{eq:sup_quality_weighting}
\end{equation}
where $\ell_i^{\mathrm{paired}}$ denotes the sample's paired-target supervision term in the equal-allocation baseline. With $t_i\in\{1,\ldots,K\}$ and the default $K=9$, the weights range from $1/9$ to $1$ and are derived from the quantized levels rather than the continuous allocation scores. All paired samples and TRFDM are retained. This control uses the level information to adjust supervision strength rather than to select the supervision endpoint.

\noindent\textbf{Default allocation-score implementation.}
We compute the default allocation score using the final output tensor of a frozen pretrained SD2.1 U-Net. Each raw image and its paired reference are independently converted to RGB, resized to $512\times512$ using bicubic interpolation, and rescaled to $[-1,1]$. The frozen VAE encodes each image using its posterior mean, multiplied by the model's latent scaling factor. The resulting latent is fed directly into the pretrained U-Net without adding noise. The time-conditioning index is fixed to $\tau_{\mathrm{score}}=999$ for all images. The pooled CLIP text embedding of an empty prompt is supplied as a single conditioning token.

Let $\phi(I)$ denote the resulting final U-Net output tensor. We apply no feature normalization before computing the mean squared discrepancy:
\begin{equation}
    \begin{aligned}
    D_i&=\operatorname{MSE}\!\left(\phi(I_i^{uw}),\phi(I_i^{la})\right)\\
    &=\frac{1}{N_{\phi}}\left\|\phi(I_i^{uw})-\phi(I_i^{la})\right\|_2^2,
    \end{aligned}
    \label{eq:supp_allocation_mse}
\end{equation}
where $N_{\phi}$ is the number of elements in each output tensor. The allocation score uses only this MSE term; cosine discrepancy and PCA-based visualization do not enter the computation. The scoring network remains frozen, whereas the enhancement U-Net is fine-tuned during LLSE training.

\noindent\textbf{IQA-rank aggregation protocol.}
For IQA-derived allocation, we measure the signed change $g_{i,m}=q_m(I_i^{la})-q_m(I_i^{uw})$ for each training pair and each of MUSIQ \cite{ke2021musiq}, URANKER \cite{guo2023uranker}, TWICEMIX \cite{fu2022twicemix}, and UIQM \cite{panetta2015uiqm}. All four metrics use a higher-is-better convention, so a larger signed change denotes a larger improvement according to that metric. We rank training pairs separately using these changes. Let $r_{i,m}^{\mathrm{train}}$ denote the resulting rank of pair $i$ under metric $m$. The four metric-wise ranks are averaged with equal weights:
\begin{equation}
    \bar r_i^{\mathrm{train}}=\frac{1}{4}\sum_{m=1}^{4}r_{i,m}^{\mathrm{train}}.
    \label{eq:sup_metric_training_ranks}
\end{equation}
The aggregate ordering is oriented so that larger estimated gains receive deeper supervision endpoints through the same level-assignment interface. The enhancement backbone, TRFDM, and remaining training settings are unchanged. This procedure aggregates ranks of input-relative score gains, rather than raw quality scores or final-reference quality rankings; it is distinct from the test-method ranks used to summarize benchmark performance.

\noindent\textbf{White-balance regularization.}
The auxiliary term denoted as WB in \cref{tab:sup_tmax_ablation} combines color and exposure regularization. During LLSE training, we decode the prediction at the assigned supervision endpoint, $z_i^{(t_i)}$, clamp the reconstructed image to $[-1,1]$, and rescale it to $[0,1]$. Bilinear resizing is applied when necessary to match the training image resolution. Let $R$, $G$, and $B$ denote the resulting channel maps, and define
\begin{equation}
    Y=0.299R+0.587G+0.114B.
    \label{eq:supp_wb_luminance}
\end{equation}
The regularizer is
\begin{equation}
    \begin{aligned}
    \mathcal L_{\mathrm{WB}}
    ={}&\left\langle(R-G)^2+(R-B)^2+(G-B)^2\right\rangle\\
    &+\left\langle[\max(Y-0.9,0)]^2\right\rangle,
    \end{aligned}
    \label{eq:supp_wb_regularizer}
\end{equation}
where the brackets denote averaging over batch and spatial dimensions. The first term penalizes differences between channels at corresponding pixel locations, while the second penalizes luminance values above $0.9$. The complete LLSE training objective is
\begin{equation}
    \begin{aligned}
    \mathcal L_{\mathrm{LLSE}}
    ={}&\operatorname{MSE}\!\left(z_i^{(t_i)},\mathcal E(I_i^{la})\right)\\
    &+0.01\,\mathcal L_{\mathrm{WB}},
    \end{aligned}
    \label{eq:supp_llse_full_objective}
\end{equation}
where the latent reconstruction term uses mean reduction over all latent elements. The auxiliary term is evaluated on the decoded LLSE prediction before TRFDM. The VAE parameters remain frozen, while gradients propagate through its decoder to the enhancement U-Net at the supervised endpoint.

For the control experiments other than the dedicated WB ablation, the UIEB-trained variants include WB, whereas the LSUI-trained variants do not. The dedicated WB comparison in \cref{tab:sup_tmax_ablation} is configured separately from these controls.

\noindent\textbf{Dataset protocol.}
The main RQUL-UIE model and its separately trained TRFDM use the complete UIEB collection for training, without an internal held-out split. The same trained model is evaluated on UIEB, EUVP, and LSUI. UIEB results characterize enhancement on the source training collection; EUVP and LSUI provide the cross-dataset comparisons. Supplementary experiments additionally use LSUI as the training source, for which EUVP and UIEB form the cross-dataset evaluation group. We retain results on all three collections and report both all-collection and cross-dataset average ranks. Training settings and evaluation resolution follow \cref{sec:exp}.

Baselines use their official checkpoints. For each table or training panel, All averages its twelve dataset--metric ranks, while Cross excludes that panel's source collection and averages the remaining eight. The same columns and comparison cohort are used for every row, including the external baseline. Source and cross-dataset roles are defined relative to our training configuration; Best-subset retains only part of the source collection.

\subsection{Choice of Allocation Proxy}
\label{sec:sup_assignment_proxy}

\begin{table*}[!t]
\centering
\setlength{\tabcolsep}{2pt}
\resizebox{\textwidth}{!}{
\begin{tabular}{l|cccc|cccc|cccc|cc}
\hlineB{2.5}
\multirow{2}{*}{Proxy / control} & \multicolumn{4}{c|}{EUVP \cite{islam2020euvp}} & \multicolumn{4}{c|}{LSUI \cite{peng2023ushape}} & \multicolumn{4}{c|}{UIEB \cite{li2019uieb}} & \multicolumn{2}{c}{Avg. rank $\downarrow$}\\
\cline{2-15}
 & MUSIQ $\uparrow$ & URANKER $\uparrow$ & TWICEMIX $\uparrow$ & UIQM $\uparrow$ & MUSIQ $\uparrow$ & URANKER $\uparrow$ & TWICEMIX $\uparrow$ & UIQM $\uparrow$ & MUSIQ $\uparrow$ & URANKER $\uparrow$ & TWICEMIX $\uparrow$ & UIQM $\uparrow$ & All & Cross \\
\hlineB{1.5}
\shortstack[l]{W/o LLSE\\(equal allocation)} & 48.462 & 2.422 & 2.256 & 2.859 & 40.357 & 2.258 & 1.522 & 2.921 & 50.692 & 2.165 & 2.664 & 3.005 & 6.08 & 6.50 \\
Random allocation & 48.487 & 2.517 & 2.126 & 2.864 & 41.153 & 2.482 & 1.587 & 2.794 & 49.511 & 2.121 & 2.353 & 2.668 & 6.17 & 5.88 \\
\hlineB{1.5}
SD2.1 (default) & 50.500 & 2.802 & 2.337 & \cellcolor{skyblue2}{3.096} & 43.223 & \cellcolor{skyblue2}{2.808} & 1.723 & 3.118 & \cellcolor{skyblue2}{51.634} & 2.408 & 2.668 & \cellcolor{skyblue1}{3.148} & 3.04 & 3.13 \\
DINOv2 \cite{oquab2023dinov2} & \cellcolor{skyblue1}{50.980} & 2.805 & \cellcolor{skyblue2}{2.380} & 3.078 & \cellcolor{skyblue1}{43.631} & 2.739 & \cellcolor{skyblue2}{1.749} & 3.116 & \cellcolor{skyblue1}{51.863} & 2.465 & \cellcolor{skyblue2}{2.737} & 3.095 & \cellcolor{skyblue2}{2.33} & \cellcolor{skyblue2}{2.38} \\
SAM2 \cite{ravi2025sam2} & \cellcolor{skyblue2}{50.605} & \cellcolor{skyblue2}{2.839} & 2.371 & 3.067 & 43.015 & 2.735 & 1.721 & \cellcolor{skyblue2}{3.125} & \cellcolor{skyblue2}{51.634} & \cellcolor{skyblue1}{2.481} & 2.715 & 3.082 & 3.04 & 3.25 \\
\shortstack[l]{IQA-rank\\aggregation} & 50.474 & \cellcolor{skyblue1}{2.879} & \cellcolor{skyblue1}{2.433} & \cellcolor{skyblue1}{3.102} & 43.229 & \cellcolor{skyblue1}{2.833} & \cellcolor{skyblue1}{1.784} & \cellcolor{skyblue1}{3.139} & 51.549 & \cellcolor{skyblue2}{2.475} & \cellcolor{skyblue1}{2.799} & \cellcolor{skyblue2}{3.124} & \cellcolor{skyblue1}{1.83} & \cellcolor{skyblue1}{1.63} \\
\hlineB{1.5}
\shortstack[l]{Semi-UIR \cite{huang2023semiuir}\\(external baseline)} & 50.410 & 2.472 & 2.213 & 2.906 & \cellcolor{skyblue2}{43.268} & 2.365 & 1.501 & 2.922 & 49.926 & 2.108 & 2.600 & 3.036 & 5.50 & 5.25 \\
\hlineB{2.5}
\end{tabular}
}
\caption{Allocation scores and controls under UIEB training. The four score-based variants share the SD2.1 enhancement backbone, TRFDM, and level-assignment procedure. Equal and random allocation reproduce the controls in \cref{tab:supervision_allocation}; Semi-UIR uses its official checkpoint. UIEB is the source collection for our configurations. All averages twelve metric ranks, and Cross averages the eight EUVP/LSUI ranks, using the same seven-row comparison cohort for every column. Raw metric values are shown; higher scores and lower average ranks are better. The \colorbox{skyblue1}{best} and \colorbox{skyblue2}{second-best} results are highlighted.}
\label{tab:assignment_proxies}
\end{table*}

\Cref{tab:assignment_proxies} compares three feature-based allocation scores and IQA-rank aggregation, with random and equal allocation as controls. All four score-based variants improve on these controls across the eight cross-dataset metrics. Their raw scores are generally close: in most cases, the spread among scoring variants is smaller than their separation from the stronger control, with a smaller gap for EUVP's TWICEMIX. IQA-rank aggregation achieves the best average rank under both All and Cross. It uses signed input-to-reference metric-score gains, whereas the default score measures changes in pretrained features. These results support several practical allocation signals for organizing input-relative correction extent within the curated collection.

\subsection{Core Ablations Across Training Datasets}
\label{sec:sup_dataset_refinement}

\begin{table*}[htbp]
\centering
\setlength{\tabcolsep}{2pt}
\resizebox{\textwidth}{!}{
\begin{tabular}{l|cccc|cccc|cccc|cc}
\hlineB{2.5}
\multirow{2}{*}{Configuration} & \multicolumn{4}{c|}{EUVP \cite{islam2020euvp}} & \multicolumn{4}{c|}{LSUI \cite{peng2023ushape}} & \multicolumn{4}{c|}{UIEB \cite{li2019uieb}} & \multicolumn{2}{c}{Avg. rank $\downarrow$}\\
\cline{2-15}
 & MUSIQ $\uparrow$ & URANKER $\uparrow$ & TWICEMIX $\uparrow$ & UIQM $\uparrow$ & MUSIQ $\uparrow$ & URANKER $\uparrow$ & TWICEMIX $\uparrow$ & UIQM $\uparrow$ & MUSIQ $\uparrow$ & URANKER $\uparrow$ & TWICEMIX $\uparrow$ & UIQM $\uparrow$ & All & Cross \\
\hlineB{1.5}
\rowcolor{gray!10} \multicolumn{15}{l}{\textit{Panel A: UIEB training source}} \\ \hline
Best-subset & \cellcolor{skyblue2}{49.890 (2)} & 2.553 (3) & 2.309 (3) & 3.022 (3) & 41.570 (3) & 2.403 (3) & 1.657 (3) & 3.023 (3) & \cellcolor{skyblue2}{51.443 (2)} & \cellcolor{skyblue2}{2.345 (2)} & \cellcolor{skyblue1}{2.753 (1)} & 3.113 (3) & 2.58 & 2.88 \\
W/o TRFDM & 47.269 (4) & \cellcolor{skyblue2}{2.775 (2)} & \cellcolor{skyblue1}{2.525 (1)} & \cellcolor{skyblue1}{3.316 (1)} & \cellcolor{skyblue2}{42.326 (2)} & \cellcolor{skyblue2}{2.789 (2)} & \cellcolor{skyblue1}{1.961 (1)} & \cellcolor{skyblue1}{3.298 (1)} & 47.236 (4) & 2.322 (3) & \cellcolor{skyblue2}{2.672 (2)} & \cellcolor{skyblue1}{3.249 (1)} & \cellcolor{skyblue2}{2.00} & \cellcolor{skyblue2}{1.75} \\
\shortstack[l]{W/o LLSE\\(equal allocation)} & 48.462 (3) & 2.422 (4) & 2.256 (4) & 2.859 (4) & 40.357 (4) & 2.258 (4) & 1.522 (4) & 2.921 (4) & 50.692 (3) & 2.165 (4) & 2.664 (4) & 3.005 (4) & 3.83 & 3.88 \\
Full model & \cellcolor{skyblue1}{50.500 (1)} & \cellcolor{skyblue1}{2.802 (1)} & \cellcolor{skyblue2}{2.337 (2)} & \cellcolor{skyblue2}{3.096 (2)} & \cellcolor{skyblue1}{43.223 (1)} & \cellcolor{skyblue1}{2.808 (1)} & \cellcolor{skyblue2}{1.723 (2)} & \cellcolor{skyblue2}{3.118 (2)} & \cellcolor{skyblue1}{51.634 (1)} & \cellcolor{skyblue1}{2.408 (1)} & 2.668 (3) & \cellcolor{skyblue2}{3.148 (2)} & \cellcolor{skyblue1}{1.58} & \cellcolor{skyblue1}{1.50} \\
\hlineB{1.5}
\rowcolor{gray!10} \multicolumn{15}{l}{\textit{Panel B: LSUI training source}} \\ \hline
Best-subset & 50.136 (4) & 1.781 (4) & 1.761 (4) & 3.002 (3) & 42.260 (4) & 1.611 (3) & 1.221 (4) & 2.978 (3) & \cellcolor{skyblue2}{51.133 (2)} & 1.756 (4) & \cellcolor{skyblue2}{2.156 (2.5)} & \cellcolor{skyblue2}{3.018 (2)} & 3.29 & 3.19 \\
W/o TRFDM & \cellcolor{skyblue1}{52.339 (1)} & \cellcolor{skyblue2}{1.957 (2)} & \cellcolor{skyblue1}{2.050 (1)} & \cellcolor{skyblue2}{3.044 (2)} & \cellcolor{skyblue1}{46.817 (1)} & \cellcolor{skyblue2}{1.767 (2)} & \cellcolor{skyblue1}{1.512 (1)} & \cellcolor{skyblue2}{3.005 (2)} & 47.546 (4) & 1.886 (3) & 1.924 (4) & 2.985 (4) & \cellcolor{skyblue2}{2.25} & \cellcolor{skyblue2}{2.63} \\
\shortstack[l]{W/o LLSE\\(equal allocation)} & 51.109 (3) & 1.875 (3) & \cellcolor{skyblue2}{2.023 (2)} & 2.912 (4) & 43.383 (3) & 1.592 (4) & \cellcolor{skyblue2}{1.274 (2)} & 2.945 (4) & 50.774 (3) & \cellcolor{skyblue2}{1.893 (2)} & \cellcolor{skyblue1}{2.446 (1)} & 2.996 (3) & 2.83 & \cellcolor{skyblue2}{2.63} \\
Full model & \cellcolor{skyblue2}{52.128 (2)} & \cellcolor{skyblue1}{1.971 (1)} & 1.799 (3) & \cellcolor{skyblue1}{3.109 (1)} & \cellcolor{skyblue2}{44.363 (2)} & \cellcolor{skyblue1}{1.804 (1)} & 1.262 (3) & \cellcolor{skyblue1}{3.047 (1)} & \cellcolor{skyblue1}{51.788 (1)} & \cellcolor{skyblue1}{1.934 (1)} & \cellcolor{skyblue2}{2.156 (2.5)} & \cellcolor{skyblue1}{3.089 (1)} & \cellcolor{skyblue1}{1.63} & \cellcolor{skyblue1}{1.56} \\
\hlineB{2.5}
\end{tabular}
}
\caption{Core ablations grouped by training source. Best-subset retains the highest allocation-score bin (\texttt{discard9}); W/o LLSE uses a common final endpoint with TRFDM retained; W/o TRFDM omits refinement. UIEB-trained entries reproduce \cref{tab:supervision_allocation}. Parentheses show ranks within each four-row panel. All averages its twelve metric ranks; Cross excludes that panel's source collection and averages EUVP/LSUI for UIEB training or EUVP/UIEB for LSUI training. Each panel uses the same columns and comparison cohort for every row. The \colorbox{skyblue1}{best} and \colorbox{skyblue2}{second-best} results are highlighted within each panel.}
\label{tab:dataset_refinement_ablation}
\end{table*}

\Cref{tab:dataset_refinement_ablation} compares the same core variants under UIEB and LSUI training. On the cross-dataset collections, the UIEB-trained full model improves over equal allocation across all four metrics. With LSUI training, MUSIQ, URANKER, and UIQM favor allocated supervision, while TWICEMIX favors equal allocation. Retaining all pairs with endpoint allocation also outperforms Best-subset under both All and Cross for each training source. The comparisons favor organizing the corrections represented in both curated reference sets, with metric-dependent differences across training settings.

The effect of texture refinement also depends on the training source. With UIEB training, TRFDM improves MUSIQ and URANKER on EUVP/LSUI, while TWICEMIX and UIQM favor the coarse output. With LSUI training, refinement improves URANKER and UIQM on EUVP/UIEB; MUSIQ and TWICEMIX decrease on EUVP but improve on UIEB. The example in \cref{fig:texturefromraw_ab} shows clearer local textures and edges after refinement, complementing these metric-dependent observations.

\subsection{Maximum Recurrent Depth \texorpdfstring{$K$}{K}}
\label{sec:sup_tmax_ablation}

\begin{table*}[htbp]
\centering
\setlength{\tabcolsep}{2pt}
\resizebox{\textwidth}{!}{
\begin{tabular}{l|cccc|cccc|cccc|cc}
\hlineB{2.5}
\multirow{2}{*}{Ablations} & \multicolumn{4}{c|}{EUVP \cite{islam2020euvp}} & \multicolumn{4}{c|}{LSUI \cite{peng2023ushape}} & \multicolumn{4}{c|}{UIEB \cite{li2019uieb}} & \multicolumn{2}{c}{Avg. rank $\downarrow$}\\
\cline{2-15}
 & MUSIQ $\uparrow$ & URANKER $\uparrow$ & TWICEMIX $\uparrow$ & UIQM $\uparrow$ & MUSIQ $\uparrow$ & URANKER $\uparrow$ & TWICEMIX $\uparrow$ & UIQM $\uparrow$ & MUSIQ $\uparrow$ & URANKER $\uparrow$ & TWICEMIX $\uparrow$ & UIQM $\uparrow$ & All & Cross \\
\hlineB{1.5}
\rowcolor{gray!10} \multicolumn{15}{l}{\textit{UIEB training source}} \\ \hline
$K=3$ & \cellcolor{skyblue1}{50.648 (1)} & 2.589 (6) & \cellcolor{skyblue1}{2.340 (1)} & 3.049 (5) & 42.405 (6) & 2.420 (6) & 1.678 (6) & 3.079 (6) & 51.471 (4) & 2.250 (6) & \cellcolor{skyblue2}{2.655 (2.5)} & 3.106 (6) & 4.63 & 4.63 \\
$K=6$ & 50.367 (5) & 2.723 (5) & 2.332 (4) & 3.048 (6) & 42.796 (5) & 2.651 (5) & 1.683 (4) & 3.081 (5) & \cellcolor{skyblue2}{51.596 (2)} & 2.387 (3) & \cellcolor{skyblue2}{2.655 (2.5)} & 3.122 (5) & 4.29 & 4.88 \\
\shortstack[l]{$K=9$,\\w/o WB} & \cellcolor{skyblue2}{50.551 (2)} & 2.758 (4) & 2.334 (3) & 3.096 (3.5) & 43.354 (3) & 2.722 (4) & 1.701 (3) & \cellcolor{skyblue2}{3.132 (2)} & 51.546 (3) & 2.382 (4) & 2.633 (4) & 3.153 (3) & 3.21 & 3.06 \\
\shortstack[l]{Full model\\($K=9$)} & 50.500 (3) & \cellcolor{skyblue2}{2.802 (2)} & \cellcolor{skyblue2}{2.337 (2)} & 3.096 (3.5) & 43.223 (4) & \cellcolor{skyblue2}{2.808 (2)} & \cellcolor{skyblue2}{1.723 (2)} & 3.118 (3) & \cellcolor{skyblue1}{51.634 (1)} & \cellcolor{skyblue2}{2.408 (2)} & \cellcolor{skyblue1}{2.668 (1)} & 3.148 (4) & \cellcolor{skyblue1}{2.46} & \cellcolor{skyblue2}{2.69} \\
$K=12$ & 50.425 (4) & \cellcolor{skyblue1}{2.830 (1)} & 2.304 (5) & \cellcolor{skyblue2}{3.101 (2)} & \cellcolor{skyblue1}{43.619 (1)} & \cellcolor{skyblue1}{2.877 (1)} & \cellcolor{skyblue1}{1.738 (1)} & 3.105 (4) & 51.309 (5) & \cellcolor{skyblue1}{2.417 (1)} & 2.587 (5) & \cellcolor{skyblue2}{3.159 (2)} & \cellcolor{skyblue2}{2.67} & \cellcolor{skyblue1}{2.38} \\
$K=15$ & 50.063 (6) & 2.764 (3) & 2.274 (6) & \cellcolor{skyblue1}{3.144 (1)} & \cellcolor{skyblue2}{43.388 (2)} & 2.758 (3) & 1.681 (5) & \cellcolor{skyblue1}{3.160 (1)} & 51.192 (6) & 2.362 (5) & 2.558 (6) & \cellcolor{skyblue1}{3.178 (1)} & 3.75 & 3.38 \\
\hlineB{1.5}
\rowcolor{gray!10} \multicolumn{15}{l}{\textit{LSUI training source}} \\ \hline
$K=3$ & 51.083 (6) & 1.936 (6) & \cellcolor{skyblue1}{1.982 (1)} & 3.077 (6) & 43.252 (6) & 1.693 (6) & \cellcolor{skyblue2}{1.295 (2)} & 3.015 (6) & 50.903 (6) & 1.876 (6) & \cellcolor{skyblue1}{2.370 (1)} & 3.091 (4) & 4.67 & 4.50 \\
$K=6$ & 51.455 (5) & 1.955 (5) & \cellcolor{skyblue2}{1.874 (2)} & 3.089 (5) & 43.605 (5) & 1.738 (5) & 1.278 (4) & 3.026 (5) & 51.203 (5) & 1.897 (3.5) & \cellcolor{skyblue2}{2.252 (2)} & 3.072 (6) & 4.38 & 4.19 \\
\shortstack[l]{$K=9$,\\w/o WB} & 51.794 (3) & 1.992 (3) & 1.835 (3) & 3.119 (3) & 44.228 (3) & 1.830 (3) & 1.287 (3) & 3.050 (3) & 51.416 (4) & \cellcolor{skyblue2}{1.905 (2)} & 2.175 (3) & 3.098 (3) & 3.00 & \cellcolor{skyblue2}{3.00} \\
\shortstack[l]{Full model\\($K=9$)} & \cellcolor{skyblue1}{52.128 (1)} & 1.971 (4) & 1.799 (5) & 3.109 (4) & \cellcolor{skyblue1}{44.363 (1)} & 1.804 (4) & 1.262 (5) & 3.047 (4) & \cellcolor{skyblue1}{51.788 (1)} & \cellcolor{skyblue1}{1.934 (1)} & 2.156 (4) & 3.089 (5) & 3.25 & 3.13 \\
$K=12$ & \cellcolor{skyblue2}{52.060 (2)} & \cellcolor{skyblue2}{2.010 (2)} & 1.807 (4) & \cellcolor{skyblue2}{3.127 (2)} & \cellcolor{skyblue2}{44.302 (2)} & \cellcolor{skyblue2}{1.837 (2)} & 1.258 (6) & \cellcolor{skyblue2}{3.057 (2)} & \cellcolor{skyblue2}{51.654 (2)} & 1.897 (3.5) & 2.099 (5) & \cellcolor{skyblue2}{3.102 (2)} & \cellcolor{skyblue2}{2.88} & \cellcolor{skyblue1}{2.81} \\
$K=15$ & 51.767 (4) & \cellcolor{skyblue1}{2.029 (1)} & 1.784 (6) & \cellcolor{skyblue1}{3.130 (1)} & 44.118 (4) & \cellcolor{skyblue1}{1.907 (1)} & \cellcolor{skyblue1}{1.302 (1)} & \cellcolor{skyblue1}{3.059 (1)} & 51.472 (3) & 1.890 (5) & 2.060 (6) & \cellcolor{skyblue1}{3.110 (1)} & \cellcolor{skyblue1}{2.83} & 3.38 \\
\hlineB{2.5}
\end{tabular}
}
\caption{Maximum-depth ablations. $K$ also sets the number of supervision levels; the full model uses $K=9$. The $K=9$, w/o WB row reports the white-balance ablation. Parentheses show ranks within each six-row training group. All averages its twelve metric ranks; Cross excludes that group's source collection and averages EUVP/LSUI for UIEB training or EUVP/UIEB for LSUI training, using the same comparison cohort. The \colorbox{skyblue1}{best} and \colorbox{skyblue2}{second-best} results are highlighted within each group.}
\label{tab:sup_tmax_ablation}
\end{table*}

\Cref{tab:sup_tmax_ablation} varies the maximum depth $K$, jointly changing the number of supervision levels, the maximum recurrent depth, and the spacing of inherited time indices. The preferred setting depends on the aggregation scope. The all-collection averages favor the default depth under UIEB training and $K=15$ under LSUI training, whereas the cross-dataset averages favor $K=12$ in both cases. We retain $K=9$ as the default configuration used for the main comparisons. White-balance regularization also has a training-dependent effect: at the default depth, it improves the reported aggregates under UIEB training but not under LSUI training.

\subsection{Sample Filtering by Assignment Scores}
\label{sec:sup_discard_ablation}

\begin{table*}[htbp]
\centering
\setlength{\tabcolsep}{2pt}
\resizebox{\textwidth}{!}{
\begin{tabular}{l|cccc|cccc|cccc|cc}
\hlineB{2.5}
\multirow{2}{*}{Ablations} & \multicolumn{4}{c|}{EUVP \cite{islam2020euvp}} & \multicolumn{4}{c|}{LSUI \cite{peng2023ushape}} & \multicolumn{4}{c|}{UIEB \cite{li2019uieb}} & \multicolumn{2}{c}{Avg. rank $\downarrow$}\\
\cline{2-15}
 & MUSIQ $\uparrow$ & URANKER $\uparrow$ & TWICEMIX $\uparrow$ & UIQM $\uparrow$ & MUSIQ $\uparrow$ & URANKER $\uparrow$ & TWICEMIX $\uparrow$ & UIQM $\uparrow$ & MUSIQ $\uparrow$ & URANKER $\uparrow$ & TWICEMIX $\uparrow$ & UIQM $\uparrow$ & All & Cross \\
\hlineB{1.5}
\rowcolor{gray!10} \multicolumn{15}{l}{\textit{UIEB training source}} \\ \hline
discard3 & 50.014 (3) & 2.393 (4) & 2.267 (3) & 2.932 (4) & 41.558 (4) & 2.219 (4) & 1.586 (3) & 2.968 (4) & 51.267 (4) & 2.137 (4) & 2.630 (4) & 3.022 (4) & 3.75 & 3.63 \\
discard6 & \cellcolor{skyblue2}{50.058 (2)} & 2.460 (3) & 2.248 (4) & 2.964 (3) & \cellcolor{skyblue2}{41.804 (2)} & 2.305 (3) & 1.585 (4) & 2.987 (3) & 51.302 (3) & 2.222 (3) & \cellcolor{skyblue2}{2.669 (2)} & 3.066 (3) & 2.92 & 3.00 \\
discard9 & 49.890 (4) & \cellcolor{skyblue2}{2.553 (2)} & \cellcolor{skyblue2}{2.309 (2)} & \cellcolor{skyblue2}{3.022 (2)} & 41.570 (3) & \cellcolor{skyblue2}{2.403 (2)} & \cellcolor{skyblue2}{1.657 (2)} & \cellcolor{skyblue2}{3.023 (2)} & \cellcolor{skyblue2}{51.443 (2)} & \cellcolor{skyblue2}{2.345 (2)} & \cellcolor{skyblue1}{2.753 (1)} & \cellcolor{skyblue2}{3.113 (2)} & \cellcolor{skyblue2}{2.17} & \cellcolor{skyblue2}{2.38} \\
Full model & \cellcolor{skyblue1}{50.500 (1)} & \cellcolor{skyblue1}{2.802 (1)} & \cellcolor{skyblue1}{2.337 (1)} & \cellcolor{skyblue1}{3.096 (1)} & \cellcolor{skyblue1}{43.223 (1)} & \cellcolor{skyblue1}{2.808 (1)} & \cellcolor{skyblue1}{1.723 (1)} & \cellcolor{skyblue1}{3.118 (1)} & \cellcolor{skyblue1}{51.634 (1)} & \cellcolor{skyblue1}{2.408 (1)} & 2.668 (3) & \cellcolor{skyblue1}{3.148 (1)} & \cellcolor{skyblue1}{1.17} & \cellcolor{skyblue1}{1.00} \\
\hlineB{1.5}
\rowcolor{gray!10} \multicolumn{15}{l}{\textit{LSUI training source}} \\ \hline
discard3 & \cellcolor{skyblue2}{51.140 (2)} & 1.911 (3) & \cellcolor{skyblue1}{2.021 (1)} & 2.943 (4) & \cellcolor{skyblue2}{43.515 (2)} & 1.629 (3) & \cellcolor{skyblue2}{1.287 (2)} & 2.952 (4) & 50.807 (3) & \cellcolor{skyblue2}{1.921 (2)} & \cellcolor{skyblue1}{2.454 (1)} & 3.006 (4) & 2.58 & \cellcolor{skyblue2}{2.50} \\
discard6 & 50.829 (3) & \cellcolor{skyblue2}{1.916 (2)} & \cellcolor{skyblue2}{1.999 (2)} & 2.992 (3) & 43.266 (3) & \cellcolor{skyblue2}{1.660 (2)} & \cellcolor{skyblue1}{1.295 (1)} & \cellcolor{skyblue2}{2.989 (2)} & 50.733 (4) & 1.886 (3) & \cellcolor{skyblue2}{2.382 (2)} & 3.015 (3) & \cellcolor{skyblue2}{2.50} & 2.75 \\
discard9 & 50.136 (4) & 1.781 (4) & 1.761 (4) & \cellcolor{skyblue2}{3.002 (2)} & 42.260 (4) & 1.611 (4) & 1.221 (4) & 2.978 (3) & \cellcolor{skyblue2}{51.133 (2)} & 1.756 (4) & 2.156 (3.5) & \cellcolor{skyblue2}{3.018 (2)} & 3.38 & 3.19 \\
Full model & \cellcolor{skyblue1}{52.128 (1)} & \cellcolor{skyblue1}{1.971 (1)} & 1.799 (3) & \cellcolor{skyblue1}{3.109 (1)} & \cellcolor{skyblue1}{44.363 (1)} & \cellcolor{skyblue1}{1.804 (1)} & 1.262 (3) & \cellcolor{skyblue1}{3.047 (1)} & \cellcolor{skyblue1}{51.788 (1)} & \cellcolor{skyblue1}{1.934 (1)} & 2.156 (3.5) & \cellcolor{skyblue1}{3.089 (1)} & \cellcolor{skyblue1}{1.54} & \cellcolor{skyblue1}{1.56} \\
\hlineB{2.5}
\end{tabular}
}
\caption{Filtering by allocation scores under UIEB and LSUI training. Each \texttt{discard} number specifies the number of score groups, with only the highest-score group retained; \texttt{discard9} is Best-subset in \cref{tab:supervision_allocation,tab:dataset_refinement_ablation}. Parentheses show ranks within each four-row training group. All averages its twelve metric ranks; Cross excludes that group's source collection and averages EUVP/LSUI for UIEB training or EUVP/UIEB for LSUI training, using the same comparison cohort. The \colorbox{skyblue1}{best} and \colorbox{skyblue2}{second-best} results are highlighted within each group.}
\label{tab:sup_discard_ablation}
\end{table*}

\Cref{tab:sup_discard_ablation} compares supervision allocation using all paired samples with filtering by the same allocation scores. Each \texttt{discard} setting divides the reference set into the indicated number of score groups and retains only the highest-score group; the remaining settings follow the corresponding full model. These groups rank estimated input-to-reference correction extent rather than final-reference quality.

Retaining all pairs and allocating their supervision produces better All and Cross averages under both training sources. The filtering alternatives discard a large fraction of the available paired samples, whereas LLSE continues to use each reference at its assigned endpoint. Within these human-selected reference sets, the comparison favors organizing the available corrections over restricting training to the highest-score group.

\section{Detailed Architecture and Computational Complexity of TRFDM}
We provide the structural specifications of \texttt{TRFDM} and analytical operation-count estimates for its computational complexity. 

\subsection{Architectural Details and Parameter Count}
TRFDM uses a U-Net-like encoder-decoder with channel widths $16,32,64,128$. A $3\times3$ convolution and GELU project the six-channel input to 16 channels. Three frequency-modulation encoder blocks are each followed by $2\times2$ max pooling, and a bottleneck maps 64 channels to 128. Each decoder stage uses a $2\times2$, stride-2 transposed convolution, projects the corresponding encoder skip through a $1\times1$ convolution, concatenates the two features, and applies a frequency-modulation block. The final projection is a biased $1\times1$ convolution from 16 to three channels.

For a frequency block with $C$ input channels and $D$ output channels, the amplitude branch uses $C\!\to\!C\!\to\!C$ pointwise convolutions with GELU between them. The phase branch concatenates cosine and sine maps, then applies two linear $1\times1$ convolutions, $2C\!\to\!2C\!\to\!C$, without an intervening activation. The modulated amplitude and phase reconstruct a complex spectrum; the real inverse-FFT result is added to the block input before a $C\!\to\!D$ pointwise convolution and GELU. The implementation applies FFT/IFFT over the spatial dimensions and uses matched spectrum shifts with the default all-dimension shift setting.

The model has \textbf{290,499} parameters (\textbf{0.290499 M}), verified from its saved parameter tensors. \Cref{tab:arch_breakdown} reports exact counts, including all three skip-projection convolutions. FP32 parameter storage is 1,161,996 bytes, or approximately \textbf{1.11 MiB}; this excludes checkpoint metadata and runtime activations.

\begin{table*}[htbp]
\centering
\caption{TRFDM architecture for a $1\times6\times512\times512$ input. Exact parameter counts include biases. Decoder groups include transposed convolution, skip projection, and frequency modulation. Output shapes omit the batch dimension.}
\label{tab:arch_breakdown}
\small
\begin{tabular}{llcc}
\hline
\textbf{Module Stage} & \textbf{Sub-Component} & \textbf{Output Shape ($C \times H \times W$)} & \textbf{Params} \\ \hline
Input Projection & \texttt{conv\_in} & $16 \times 512 \times 512$ & 880 \\ \hline
Encoder Stage 1 & \texttt{enc\_blocks.0} & $16 \times 512 \times 512$ & 2,400 \\
Downsampling 1 & \texttt{pool} & $16 \times 256 \times 256$ & -- \\ \hline
Encoder Stage 2 & \texttt{enc\_blocks.1} & $32 \times 256 \times 256$ & 2,672 \\
Downsampling 2 & \texttt{pool} & $32 \times 128 \times 128$ & -- \\ \hline
Encoder Stage 3 & \texttt{enc\_blocks.2} & $64 \times 128 \times 128$ & 10,464 \\
Downsampling 3 & \texttt{pool} & $64 \times 64 \times 64$ & -- \\ \hline
Bottleneck & \texttt{bottleneck} & $128 \times 64 \times 64$ & 41,408 \\ \hline
Decoder Stage 3 & \shortstack[l]{\texttt{up\_convs.0} + \texttt{skip\_conv\_11.0}\\+ \texttt{up\_blocks.0}} & $64 \times 128 \times 128$ & 176,960 \\ \hline
Decoder Stage 2 & \shortstack[l]{\texttt{up\_convs.1} + \texttt{skip\_conv\_11.1}\\+ \texttt{up\_blocks.1}} & $32 \times 256 \times 256$ & 44,448 \\ \hline
Decoder Stage 1 & \shortstack[l]{\texttt{up\_convs.2} + \texttt{skip\_conv\_11.2}\\+ \texttt{up\_blocks.2}} & $16 \times 512 \times 512$ & 11,216 \\ \hline
Output Projection & \texttt{conv\_out} & $3 \times 512 \times 512$ & 51 \\ \hline
\textbf{Total} & & & \textbf{290,499} \\ \hline
\end{tabular}
\end{table*}

\subsection{Computational Complexity Analysis (FLOPs) and Inference Efficiency}
We count convolutional multiply--accumulates (MACs) from the implementation and supplement them with analytical estimates for the FFT and associated operations. Consistent with the convolution-counting convention of \texttt{fvcore}, one convolutional MAC counts as one operation. The frequency-domain estimates use the conventions below.

For a 2D Radix-2 FFT/IFFT operating on a feature map of size $C \times H \times W$, one complex multiplication requires 4 real multiplications and 2 real additions (6 FLOPs), while one complex addition requires 2 real additions (2 FLOPs). Under these counting conventions, the estimated operation counts are:
\begin{align}
    \text{FLOPs}_{\text{FFT2}} &= 5 \cdot C \cdot H \cdot W \cdot \log_2(H \cdot W) \\
    \text{FLOPs}_{\text{IFFT2}} &= 5 \cdot C \cdot H \cdot W \cdot \log_2(H \cdot W) + C \cdot H \cdot W \\
    \text{FLOPs}_{\text{Abs}} &= 4 \cdot C \cdot H \cdot W \quad (\text{for } \sqrt{a^2+b^2}) \\
    \text{FLOPs}_{\text{Angle}} &= 2 \cdot C \cdot H \cdot W \quad (\text{for } \arg(a+jb)) \\
    \text{FLOPs}_{\text{Recon}} &= 6 \cdot C \cdot H \cdot W \quad (\text{for } A \cdot e^{j\phi})
\end{align}
For a frequency block whose \emph{input} has dimensions $C\times H\times W$, the estimated frequency-operation count is:
\begin{equation}
    \text{FLOPs}_{total} = 10 \cdot C \cdot H \cdot W \cdot \log_2(H \cdot W) + 13 \cdot C \cdot H \cdot W
    \label{eq:sup_frequency_flops}
\end{equation}

The FFT precedes channel fusion, so its channel count is the block's input width rather than its output width. Across the seven blocks, \cref{eq:sup_frequency_flops} gives \textbf{3.7715 G} estimated frequency operations. Convolutions, including transposed convolutions and skip projections, require \textbf{8.7955 G} MACs. Under the stated counting convention, their sum is \textbf{12.5669 GFLOPs}; \cref{tab:flops_breakdown} provides the breakdown. This is an analytical counted-operation estimate, excluding biases, GELU, phase sine/cosine evaluations, residual additions, pooling, padding, and data rearrangement. The fixed costs assigned to angle extraction and complex reconstruction are approximations. The reporting resolution of $512\times512$ differs from the $256\times256$ main quality evaluation setting.
\subsection{Computational Overhead Comparison}
We compare the computational footprint of \textbf{TRFDM} with the recurrent latent enhancement backbone, comprising the VAE encoder, recurrent U-Net, and VAE decoder.

\begin{table}[htbp]
\centering
\caption{Analytical operation counts at $512\times512$ input resolution. $C_{\mathrm{FFT}}$ is the block's input channel count. Totals use unrounded counts and the stated one-operation-per-convolutional-MAC convention.}
\label{tab:flops_breakdown}
\small
\resizebox{\linewidth}{!}{
\begin{tabular}{lccc}
\hline
\textbf{Stage} & 
\textbf{Resolution} & 
\boldmath$C_{\mathrm{FFT}}$ & 
\textbf{GFLOPs (est.)} \\
\hline
Encoder 1 & $512 \times 512$ & 16 & 0.8095 \\
Encoder 2 & $256 \times 256$ & 16 & 0.1814 \\
Encoder 3 & $128 \times 128$ & 32 & 0.0802 \\
Bottleneck & $64 \times 64$ & 64 & 0.0349 \\
Decoder 3 & $128 \times 128$ & 128 & 0.3209 \\
Decoder 2 & $256 \times 256$ & 64 & 0.7256 \\
Decoder 1 & $512 \times 512$ & 32 & 1.6190 \\
\hline
\multicolumn{3}{l}{\textbf{Frequency subtotal (est.)}} & \textbf{3.7715} \\
\multicolumn{3}{l}{\textbf{Convolutions (1 MAC = 1 op)}} & \textbf{8.7955} \\
\hline
\multicolumn{3}{l}{\textbf{Counted total (est.)}} & \textbf{12.5669} \\
\hline
\end{tabular}}
\end{table}

\paragraph{Inference Efficiency.} The hardware benchmark is conducted on a single NVIDIA GPU using PyTorch's high-precision benchmarking suite with 50 warm-up runs. For an input tensor of $1 \times 6 \times 512 \times 512$, the reported per-call TRFDM measurements are:
\begin{itemize}
    \item \textbf{Peak Inference Memory}: 467.12 MB (excluding gradient graphs).
    \item \textbf{Average Inference Latency}: 29.53 ms per frame.
    \item \textbf{Throughput / Frame Rate}: 33.86 FPS.
\end{itemize}

\paragraph{Overhead Comparison with the Enhancement Backbone.} Table~\ref{tab:overhead_comparison} reports component parameters, counted operations, and latency. Backbone accounting includes one VAE encoding, $K$ U-Net evaluations, and one final VAE decoding. One TRFDM call adds approximately \textbf{0.031\%} parameters, \textbf{0.45\%} counted operations under the stated convention, and \textbf{0.98\%} reported latency. Its per-call throughput does not describe the complete recurrent framework.

\begin{table}[!ht]
\centering
\caption{Component-level architectural and computational overhead of TRFDM relative to the recurrent enhancement backbone. TRFDM costs are per evaluation.}
\label{tab:overhead_comparison}
\small
\setlength{\tabcolsep}{4pt}
\resizebox{\linewidth}{!}{
\begin{tabular}{lccc}
\hline
\textbf{Component} &
\shortstack{\textbf{Params}\\\textbf{(M)}} &
\shortstack{\textbf{Complexity}\\\textbf{(GFLOPs)}} &
\textbf{Latency} \\
\hline
VAE Encoder & 34.16 & 542.21 & -- \\
VAE Decoder & 49.49 & 1242.11 & -- \\
U-Net Backbone & 865.92 & 113.24 & -- \\
\hline
\shortstack[l]{Enhancement\\backbone} &
$\sim$949.57 &
$\sim$2803.44$^\dagger$ &
$\sim$3.0 s \\
\hline
\textbf{TRFDM (Ours)} &
\textbf{0.290499} &
\textbf{12.57} &
\textbf{29.53 ms} \\
\textit{Relative overhead} &
\textit{$\approx$ +0.031\%} &
\textit{+0.45\%} &
\textit{+0.98\%} \\
\hline
\end{tabular}
}

\vspace{2pt}
\begin{flushleft}
\footnotesize
$^\dagger$ \textit{Backbone FLOPs count one VAE encoding, $K$ U-Net evaluations, and one final VAE decoding. TRFDM complexity is the analytical estimate in \cref{tab:flops_breakdown}. Latencies are reported measurements for the backbone and an individual TRFDM evaluation.}
\end{flushleft}
\end{table}

\section{Additional Qualitative Comparisons}
\begin{figure*}
    \centering
    \includegraphics[width=0.9\linewidth,height=0.95\textheight,keepaspectratio]{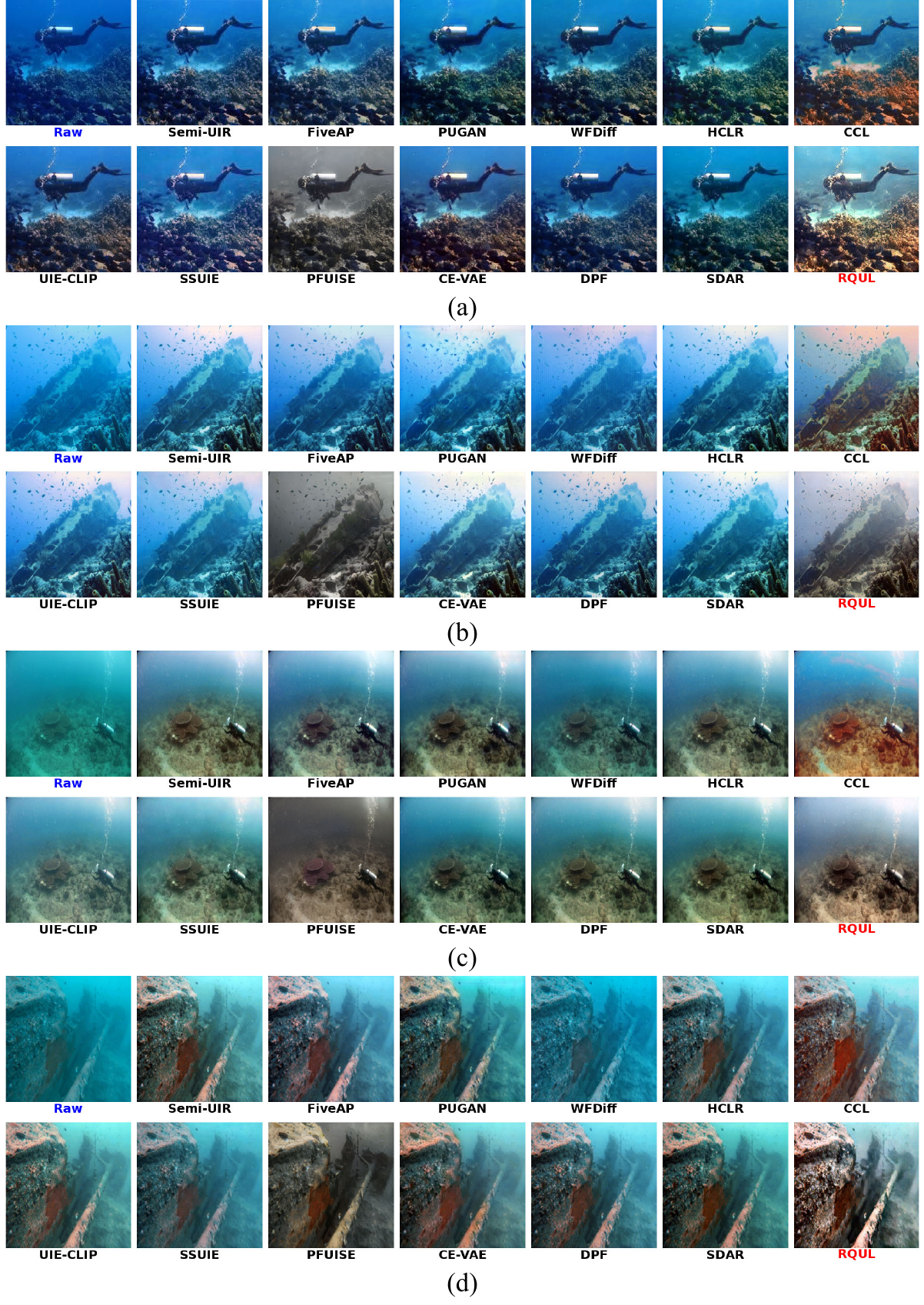}
    \caption{Visual Comparison of the enhanced results among different methods with respect to UIEB \cite{li2019uieb} dataset.}
    \label{fig:sup_uieb}
\end{figure*}

\begin{figure*}
    \centering
    \includegraphics[width=0.9\linewidth,height=0.95\textheight,keepaspectratio]{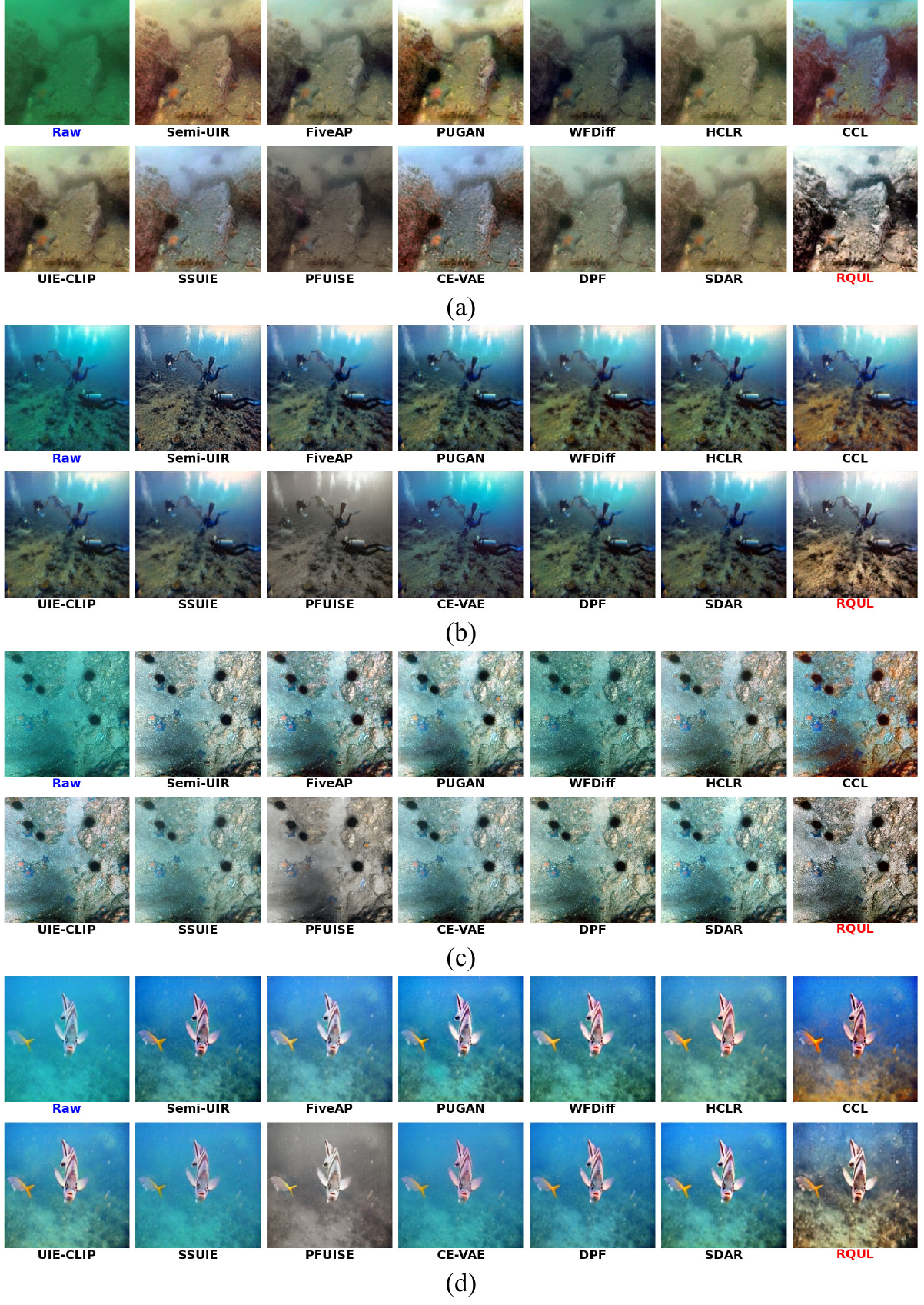}
    \caption{Visual Comparison of the enhanced results among different methods with respect to LSUI \cite{peng2023ushape} dataset.}
    \label{fig:sup_lsui}
\end{figure*}

\begin{figure*}
    \centering
    \includegraphics[width=0.9\linewidth,height=0.95\textheight,keepaspectratio]{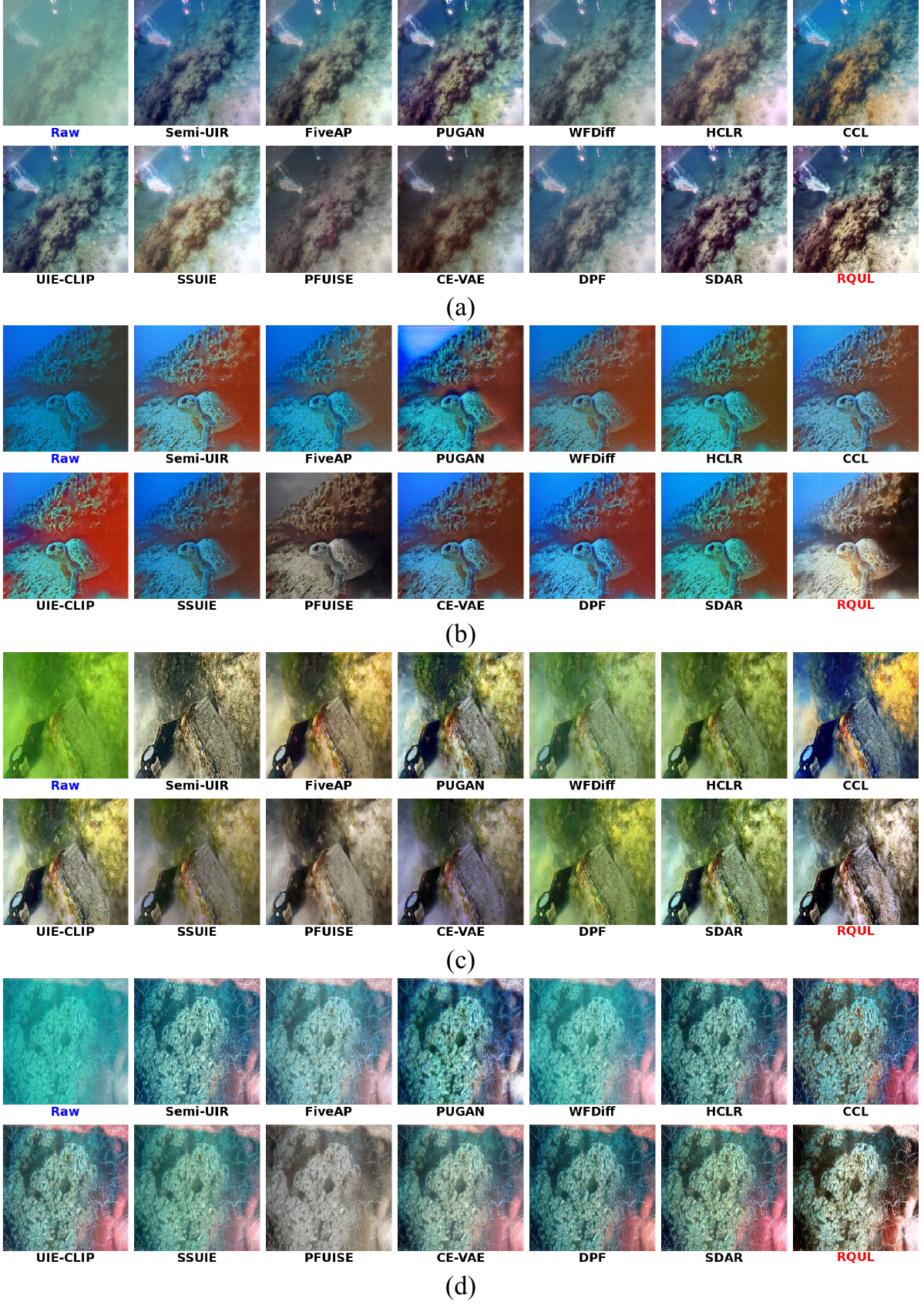}
    \caption{Visual Comparison of the enhanced results among different methods with respect to EUVP \cite{islam2020euvp} dataset.}
    \label{fig:sup_euvp}
\end{figure*}

\Cref{fig:sup_uieb,fig:sup_lsui,fig:sup_euvp} provide additional visual comparisons on UIEB, LSUI, and EUVP, respectively.

\end{document}